%% file: main.tex
\documentclass[10pt,twocolumn,letterpaper]{article}

\usepackage{cvpr}
\usepackage{times}
\usepackage{epsfig}
\usepackage{graphicx}
\usepackage{amsmath}
\usepackage{amssymb}
\usepackage{comment}
\usepackage{setspace}

\usepackage[breaklinks=true,bookmarks=false]{hyperref}

\cvprfinalcopy 

\def\cvprPaperID{2448} 

\title{On the Continuity of Rotation Representations in Neural Networks}

\author{Yi Zhou\thanks{Authors have equal contribution.}\\
University of Southern California\\
{\tt\small zhou859@usc.edu}
\and
Connelly Barnes$^*$\\
Adobe Research\\
{\tt\small connellybarnes@yahoo.com}
\and
Jingwan Lu\\
Adobe Research\\
{\tt\small jlu@adobe.com}
\and
Jimei Yang\\
Adobe Research\\
{\tt\small jimyang@adobe.com}
\and
Hao Li\\
University of Southern California, Pinscreen\\
USC Institute for Creative Technologies\\
{\tt\small hao@hao-li.com}
}

\date{November 2018}

\input{macros}

\usepackage[numbers]{natbib}
\usepackage{graphicx}
\usepackage{amssymb}
\usepackage{amsmath}
\usepackage{scalerel,stackengine}
\stackMath
\newcommand\reallywidehat[1]{%
\savestack{\tmpbox}{\stretchto{%
  \scaleto{%
    \scalerel*[\widthof{\ensuremath{#1}}]{\kern.1pt\mathchar"0362\kern.1pt}%
    {\rule{0ex}{\textheight}}
  }{\textheight}%
}{2.4ex}}%
\stackon[-6.9pt]{#1}{\tmpbox}%
}
\begin{document}

\newcommand*{\vertbar}{\rule[-1ex]{0.5pt}{2.5ex}}
\newcommand*{\horzbar}{\rule[.5ex]{2.5ex}{0.5pt}}

\maketitle

\input{abstract.tex}
\setstretch{0.91}
\input{introduction.tex}
\vspace{-3pt}
\input{related_work.tex}

\vspace{-3pt}
\section{Definition of Continuous Representation}
\label{sec:continuity_definition}

In this section, we begin by defining the terminology we will use in the paper. Next, we analyze a simple motivating example of 2D rotations. This allows us to develop our general definition of continuity of representation in neural networks. We then explain how this definition of continuity is related to concepts in topology.

\textbf{Terminology}. To denote a matrix, we typically use $M$, and $M_{ij}$ refers to its $(i,j)$ entry. We use the term $SO(n)$ to denote the \emph{special orthogonal group}, the space of $n$ dimensional rotations. This group is defined on the set of $n \times n$ real matrices with $MM^T = M^TM = I$ and $\det(M)=1$. The group operation is multiplication, which results in the concatenation of rotations. We denote the \emph{$n$ dimensional unit sphere} as $S^n = \{x \in \mathbb{R}^{n+1}: ||x||=1\}$.

\begin{figure}
    \centering
    \includegraphics[width=2.9in]{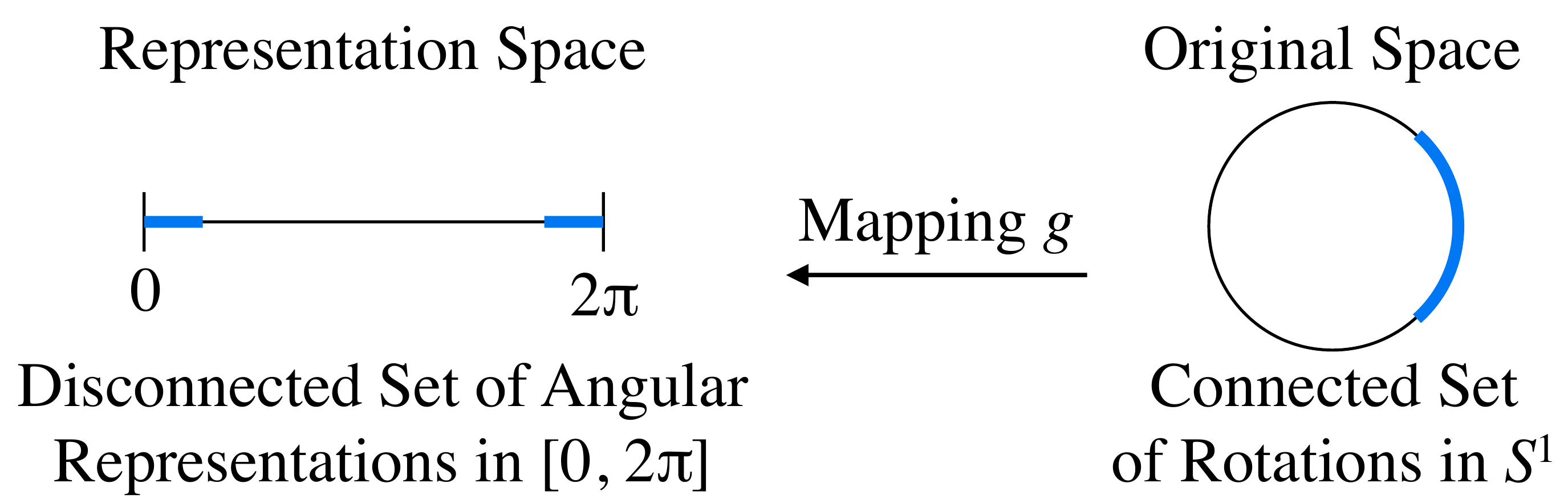}
    \caption{A simple 2D example, which motivates our definition of continuity of representation. See \sect{sec:continuity_definition} for details.}
    \label{fig:problem2d}
\end{figure}

\textbf{Motivating example: 2D rotations.} We now consider the representation of 2D rotations. For any 2D rotation $M \in SO(2)$, we can also express the matrix as:

\vspace{-6pt}
\begin{equation}
    M = \left[\begin{array}{cc}
        \cos(\theta) & -\sin(\theta) \\
        \sin(\theta) & \cos(\theta)
        \end{array}\right]
\end{equation}
\vspace{-6pt}

We can represent any rotation matrix $M \in SO(2)$ by choosing $\theta \in R$, where $R$ is a suitable set of angles, for example, $R = [0, 2\pi]$. However, this particular representation intuitively has a problem with continuity. The problem is that if we define a mapping $g$ from the original space $SO(2)$ to the angular representation space $R$, then this mapping is discontinuous. In particular, the limit of $g$ at the identity matrix, which represents zero rotation, is undefined: one directional limit gives an angle of $0$ and the other gives $2\pi$. We depict this problem visually in \fig{fig:problem2d}. On the right, we visualize a connected set of rotations $C \subset SO(2)$ by visualizing their first column vector $[\cos(\theta),\sin(\theta)]^T$ on the unit sphere $S^1$. 
On the left, after mapping them through $g$, we see that the angles are disconnected. In particular, we say that this representation is discontinuous because the mapping $g$ from the original space to the representation space is discontinuous. We argue that these kind of discontinuous representations can be harder for neural networks to fit. 
Contrarily, if we represent the 2D rotation $M \in SO(2)$ by its first column vector $[\cos(\theta),\sin(\theta)]^T$, then the representation would be continuous.

\begin{figure}
    \centering
    \includegraphics[width=3.0in]{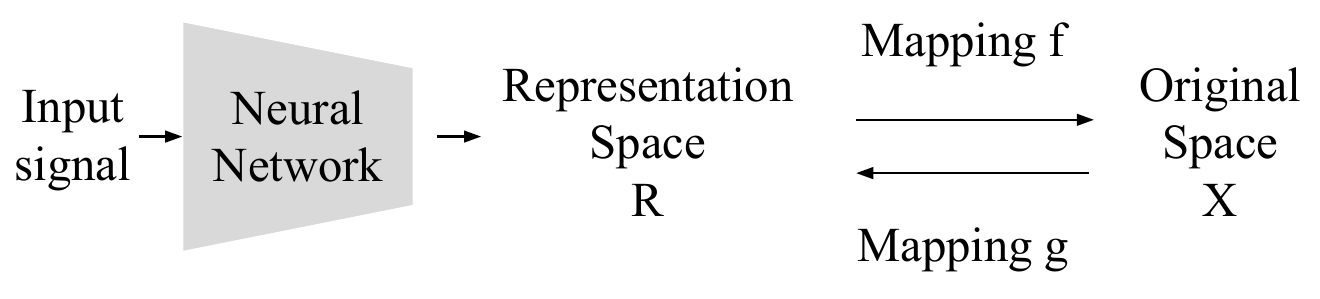}
    \caption{Our definition of continuous representation, as well as how it can apply in a neural network. See the body for details.}
    \label{fig:continuity}
\end{figure}

\textbf{Continuous representation:} We can now define what we consider a continuous representation. We illustrate our definitions graphically in \fig{fig:continuity}. Let $R$ be a subset of a real vector space equipped with the Euclidean topology. We call $R$ the \emph{representation space}: in our context, a neural network produces an intermediate representation in $R$. This neural network is depicted on the left side of \fig{fig:continuity}. We will come back to this neural network shortly. Let $X$ be a compact topological space. We call $X$ the \emph{original space}. In our context, any intermediate representation in $R$ produced by the network can be \emph{mapped} into the original space $X$. Define the \emph{mapping to the original space} $f: R \to X$, and the \emph{mapping to the representation space} $g: X \to R$. We say $(f, g)$ is a \emph{representation} if for every $x \in X, f(g(x)) = x$, that is, $f$ is a left inverse of $g$. We say the \emph{representation is continuous} if $g$ is continuous.

\textbf{Connection with neural networks:} We now return to the neural network on the left side of \fig{fig:continuity}. We imagine that inference runs from left to right. Thus, the neural network accepts some input signals on its left hand side, outputs a representation in $R$, and then passes this representation through the mapping $f$ to get an element of the original space $X$. Note that in our context, \emph{the mapping f} is implemented as a mathematical function that is used as part of the forward pass of the network at both training and inference time. Typically, at training time, we might impose losses on the original space $X$. We now describe the intuition behind why we ask that $g$ be continuous. Suppose that we have some connected set $C$ in the original space, such as the one shown on the right side of \fig{fig:problem2d}. Then if we map $C$ into representation space $R$, and $g$ is continuous, then the set $g(C)$ will remain connected. Thus, if we have continuous training data, then this will effectively create a continuous training signal for the neural network. Contrarily, if $g$ is not continuous, as shown in \fig{fig:problem2d}, then a connected set in the original space may become disconnected in the representation space. This could create a discontinuous training signal for the network. We note that the units in the neural network are typically continuous, as defined on Euclidean topology spaces. Thus, we require the representation space $R$ to have Euclidean topology because this is consistent with the continuity of the network units.

\textbf{Domain of the mapping $f$:} We additionally note that for neural networks, it is specifically beneficial for the mapping $f$ to be defined almost everywhere on a set where the neural network outputs are expected to lie. This enables $f$ to \emph{map arbitrary representations produced by the network} back to the original space $X$.

\textbf{Connection with topology:} Suppose that $(f, g)$ is a continuous representation. Note that $g$ is a continuous one-to-one function from a compact topological space to a Hausdorff space. From a theorem in  topology~\cite{kosniowski1980first}, this implies that if we restrict the codomain of $g$ to $g(X)$ (and use the subspace topology for $g(X)$) then the resulting mapping is a homeomorphism. A \emph{homeomorphism} is a continuous bijection with a continuous inverse. For geometric intuition, a homeomorphism is often described as a continuous and invertible stretching and bending of one space to another, with also a finite number of cuts allowed if one later glues back together the same points. One says that two spaces are topologically equivalent if there is a homeomorphism between them. Additionally, $g$ is a topological \emph{embedding} of the original space $X$ into the representation space $R$. Note that we also have the inverse of $g$: if we restrict $f$ to the domain $g(X)$ then the resulting function $f|_{g(X)}$ is simply the inverse of $g$. Conversely, if the original space $X$ is not homeomorphic to any subset of the representation space $R$ then there is no possible continuous representation $(f, g)$ on these spaces. We will return to this later when we show that there is no continuous representation for the 3D rotations in four or fewer dimensions. 

\vspace{-3pt}
\section{Rotation Representation Analysis}
\label{sec:continuity_rotations}
\vspace{-3pt}

Here we provide examples of rotation representations that could be used in networks. We start by looking in \sect{sec:continuity_rotations_discontinuous} at some discontinuous  representations for 3D rotations, then look in \sect{sec:continuity_rotations_continuous} at continuous rotation representations in $n$ dimensions, and show that how for the 3D rotations, these become 6D and 5D continuous rotation representations. We believe this analysis can help one to choose suitable rotation representations for learning tasks.

\subsection{Discontinuous Representations}
\label{sec:continuity_rotations_discontinuous}

\textbf{Case 1: Euler angle representation for the 3D rotations.} Let the original space $X = SO(3),$ the set of 3D rotations. Then we can easily show discontinuity in an Euler angle representation by considering the azimuth angle $\theta$ and reducing this to the motivating example for 2D rotations shown in \sect{sec:continuity_definition}. In particular, the identity rotation $I$ occurs at a discontinuity, where one directional limit gives $\theta = 0$ and the other directional limit gives $\theta = 2\pi$. We visualize the discontinuities in this representation, and all the other representations, in the supplemental \sect{sec:vis}.

\textbf{Case 2: Quaternion representation for the 3D rotations.} Define the original space $X = SO(3),$ and the representation space $Y = \mathbb{R}^4$, which we use to represent the quaternions. We can now define the mapping to the representation space {\small$g_q(M) = $
\vspace{-0.35cm}
\arraycolsep=1.4pt\def\arraystretch{1}
\begin{equation}
\begin{cases}
    \left[\begin{array}{cccc}M_{32}-M_{23}, & M_{13}-M_{31}, & M_{21}-M_{12}, & t\end{array} \right]^T & \text{if }t \neq 0 \\
    \left[\begin{array}{cccc}\sqrt{M_{11}+1}, & c_2\sqrt{M_{22}+1}, & c_3\sqrt{M_{33}+1}, & 0\end{array} \right]^T & \text{if }t = 0
\end{cases}
\label{eqn:quaternion_encoder}
\end{equation}
\begin{equation}
t = \Tr(M)+1, 
c_i = \begin{cases}
1 &  \text{  if } M_{i,1} + M_{1,i} > 0 \\
-1 & \text{  if } M_{i,1} + M_{1,i} < 0 \\
(\mathrm{sgn}(M_{3,2}))^i & \text{  Otherwise } \\
\end{cases}
\end{equation}
}

Likewise, one can define the mapping to the original space $SO(3)$ as in ~\cite{WikiQuaternion}:

\vspace{-13pt}
{
\small
\begin{multline}
f_q\left(\left[ x_0, y_0, z_0, w_0\right]\right) = \\
  \left[ {\begin{array}{ccc}
   1-2y^2-2z^2, & 2xy - 2zw, & 2xz + 2yw \\
   2xy + 2zw, & 1 - 2x^2 - 2z^2, & 2yz - 2xw \\
   2xz - 2yw, & 2yz + 2xw, & 1 - 2x^2 - 2y^2
  \end{array} } \right],\\
(x, y, z, w) = N([x_0, y_0, z_0, w_0])
\end{multline}
\label{eqn:quaternion_f}
}
Here the normalization function is defined as $N(q) = q/||q||$. By expanding in terms of the axis-angle representation for the matrix $M$, one can verify that for every $M \in SO(3), f(g(M)) = M.$

However, we find that the representation is not continuous. Geometrically, this can be seen by taking different directional limits around the matrices with 180 degree rotations, which are defined by $R_{\pi} = \{M \in SO(3): \Tr(M)=-1\}$. Specifically, in the top case of \eqn{eqn:quaternion_encoder}, where $t \ne 0$, the limit of $g_q$ as we approach a 180 degree rotation is $[\begin{smallmatrix}0,&0,&0,&0\end{smallmatrix}]$, and meanwhile, the first three coordinates of $g_q(r)$ for $r\in R_{\pi}$ are nonzero. Note that our definition of continuous representation from \sect{sec:continuity_definition}  requires a Euclidean topology for the representation space $Y$, in contrast to the usual topology for the quaternions of the real projective space $\mathbb{R}P^3,$ which we discuss in the next paragraph. In a similar way, we can show that other popular representations for the 3D rotations such as axis-angle have discontinuities, e.g. the axis in axis-angle has discontinuities at the 180 degree rotations.

\textbf{Representations for the 3D rotations are discontinuous in four or fewer dimensions.} The 3D rotation group $SO(3)$ is homeomorphic to the real projective space $\mathbb{R}P^3.$  The space $\mathbb{R}P^{n}$ is defined as the quotient space of $\mathbb{R}^{n+1} \setminus \{0\}$ under the equivalence relation that $x \sim \lambda x$ for all $\lambda \neq 0$. In a graphics and vision context, it may be most intuitive to think of $\mathbb{R}P^3$ as being the homogeneous coordinates in $\mathbb{R}^4$ with the previous equivalence relation used to construct an appropriate topology via the quotient space.

Based on standard embedding and non-embedding results in topology~\cite{davis1998embeddings}, we know that $\mathbb{R}P^3$ (and thus $SO(3)$) embeds in $\mathbb{R}^5$ with the Euclidean topology, but does not embed in $\mathbb{R}^d$ for any $d<5$. By the definition of embedding, there is no homeomorphism from $SO(3)$ to any subset of $\mathbb{R}^d$ for any $d<5$, but a continuous representation requires this. Thus, there is no such continuous representation.

\subsection{Continuous Representations}
\label{sec:continuity_rotations_continuous}

In this section, we develop two continuous representations for the $n$ dimensional rotations $SO(n)$. We then explain how for the 3D rotations $SO(3)$ these become 6D and 5D continuous rotation representations. 

\textbf{Case 3: Continuous representation with $n^2-n$ dimensions for the $n$ dimensional rotations.} The rotation representations we have considered thus far are all not continuous. One possibility to make a rotation representation continuous would be to just use the identity mapping, but this would result in matrices of size $n\times n$ for the representation, which can be excessive, and would still require orthogonalization, such as a Gram-Schmidt process in the mapping $f$ to the original space, if we want to ensure that network outputs end up back in $SO(n)$. Based on this observation, we propose to perform an orthogonalization process in the representation itself. Let the original space $X = SO(n)$, and the representation space be $R = \mathbb{R}^{n\times (n-1)} \setminus D$ ($D$ will be defined shortly). Then we can define a mapping $g_{\mathrm{GS}}$ to the representation space that simply drops the last column vector of the input matrix:

\vspace{-12pt}
{\small
\begin{equation}
g_{\mathrm{GS}}\left(\left[ {\begin{array}{ccc}
   \vertbar &  & \vertbar \\
   a_1 & \ldots & a_n \\
   \vertbar &  & \vertbar
  \end{array} } \right]\right) = 
  \left[ {\begin{array}{ccc}
   \vertbar &  & \vertbar \\
   a_1 & \ldots & a_{n-1} \\
   \vertbar &  & \vertbar
  \end{array} } \right]
\label{eqn:gramschmidt_repr}
\end{equation}
}
, where $a_i, i=1,2,...,n$ are column vectors.
We note that the set $g_{\mathrm{GS}}(X)$ is a Stiefel manifold~\cite{WikiStiefel}.  Now for the mapping $f_{\mathrm{GS}}$ to the original space, we can define the following Gram-Schmidt-like process:
{\small
\begin{equation}
f_{\mathrm{GS}}\left(\left[ {\begin{array}{ccc}
   \vertbar &  & \vertbar \\
   a_1 & \ldots & a_{n-1} \\
   \vertbar &  & \vertbar
  \end{array} } \right]\right) = 
  \left[ {\begin{array}{ccc}
   \vertbar &  & \vertbar \\
   b_1 & \ldots & b_n \\
   \vertbar &  & \vertbar
  \end{array} } \right]
\label{eqn:gramschmidt_forward_first}
\end{equation}
\begin{equation}
b_i = \left[ \begin{cases}
N(a_1) & \text{if } $i=1$ \\
N(a_i - \sum_{j=1}^{i-1}(b_{j} \boldsymbol{\cdot} a_i) b_{j}) & \text{if } 2 \leq i < n \\
\det \left[ {\begin{array}{cccc}
   \vertbar &  & \vertbar & e_1 \\
   b_1 & \ldots & b_{n-1} & \vdots \\
   \vertbar &  & \vertbar & e_n
  \end{array} } \right] & \text{if } i = n.
\end{cases}
\right]^T
\label{eqn:gramschmidt_forward_repr}
\end{equation}
}
Here $N(\cdot)$ denotes a normalization function, the same as before, and $e_1, \ldots, e_n$ are the $n$ canonical basis vectors of the Euclidean space. The only difference of $f_{\mathrm{GS}}$ from an ordinary Gram-Schmidt process is that the last column is computed by a generalization of the cross product to $n$ dimensions. Now clearly, $g_{\mathrm{GS}}$ is continuous. To check that for every $M \in SO(n)$, $f_{\mathrm{GS}}(g_{\mathrm{GS}}(M)) = M$, we can use induction and the properties of the orthonormal basis vectors in the columns of $M$ to show that the Gram-Schmidt process does not modify the first $n-1$ components. Lastly, we can use theorems for the generalized cross product such as Theorem 5.14.7 of Bloom~\cite{bloom1979linear}, to show that the last component of $f_{\mathrm{GS}}(g_{\mathrm{GS}}(M))$ agrees with $M$. Finally, we can define the set $D$ as that where the above Gram-Schmidt-like process does not map back to $SO(n)$: specifically, this is where the dimension of the span of the $n-1$ vectors input to $g_{\mathrm{GS}}$ is less than $n-1$. 

\textbf{6D representation for the 3D rotations:} For the 3D rotations, Case 3 gives us a 6D representation. The generalized cross product for $b_n$ in \eqn{eqn:gramschmidt_forward_repr} simply reduces to the ordinary cross product $b_1 \times b_2$. We give the detailed equations in \sect{sec:sixd_repr} in the supplemental document. We specifically note that using our 6D representation in a network can be beneficial because the mapping $f_{\mathrm{GS}}$ in \eqn{eqn:gramschmidt_forward_repr} ensures that the resulting 3x3 matrix is orthogonal. In contrast, suppose a direct prediction for 3x3 matrices is used. Then either the orthogonalization can be done in-network or as a postprocess. If orthogonalization is done in network, the last 3 components of the matrix will be discarded by the Gram-Schmidt process in \eqn{eqn:gramschmidt_forward_repr}, so the 3x3 matrix representation is effectively our 6D representation plus 3 useless parameters. If orthogonalization is done as a postprocess, then this prevents certain applications such as forward kinematics, and the error is also higher as shown in \sect{sec:empirical_results}.

\textbf{Group operations such as multiplication:} Suppose that the original space is a group such as the rotation group, and we want to multiply two representations $r_1,r_2 \in R$. In general, we can do this by first mapping to the original space, multiplying the two elements, and then mapping back: $r_1r_2  = g(f(r_1)f(r_2)).$ However, for the proposed representation here, we can gain some computational efficiency as follows. Since the mapping to the representation space in \eqn{eqn:gramschmidt_repr} drops the last column, when computing $f(r_2)$, we can simply drop the last column and compute the product representation as the product of an $n \times n$ and an $n \times (n-1)$ matrix.

\begin{figure}
    \centering
    \includegraphics[width=1.2in]{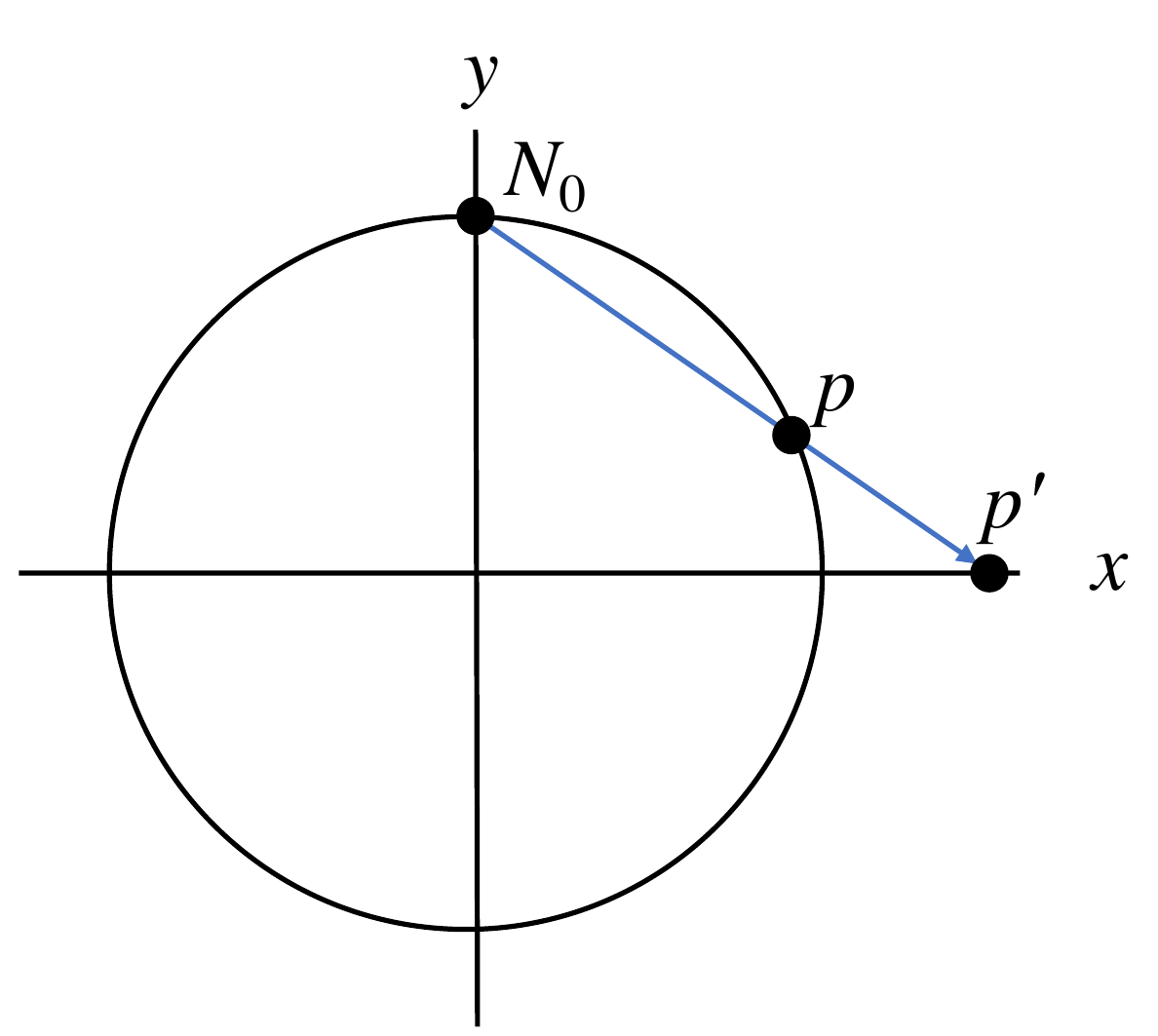}
    \caption{An illustration of stereographic projection in 2D. We are given as input a point $p$ on the unit sphere $S^1$. We construct a ray from a fixed projection point $N_0 = (0, 1)$ through $p$ and find the intersection of this ray with the plane $y = 0$. The resulting point $p'$ is the stereographic projection of $p$.}
    \vspace{-0.4cm}
    \label{fig:stereographic}
\end{figure}

\textbf{Case 4: Further reducing the dimensionality for the $n$ dimensional rotations.} For $n \ge 3$ dimensions, we can reduce the dimension for the representation in the previous case, while still keeping a continuous representation. Intuitively, a lower dimensional representation that is less redundant could be easier to learn. However, we found in our experiments that the dimension-reduced representation does not outperform the Gram-Schmidt-like representation from Case 3. However, we still develop this representation because it allows us to show that continuous rotation representations can outperform discontinuous ones. 

We show that we can perform such dimension reduction using one or more stereographic projections combined with normalization, similar to Hopf~\cite{hopf1940systeme}. We show an illustration of a 2D stereographic projection in \fig{fig:stereographic}, which can be easily generalized to higher dimensions. Let us first normalize the input point, so it projects to a sphere, and then stereographically project the result using a projection point of $(1, 0, \ldots, 0)$. We call this combined operation a \emph{normalized projection,} and define it as $P: \mathbb{R}^m \to \mathbb{R}^{m-1}$:
\begin{equation}
    P(u) = \left[\begin{array}{cccc}
        \frac{v_2}{1-v_1}, & \frac{v_3}{1-v_1}, & \ldots, & \frac{v_{m}}{1-v_1}
    \end{array}\right]^T\text{, } v = u/||u||.
\label{eqn:normalized_project}
\end{equation}
Now define a function $Q: \mathbb{R}^{m-1} \to \mathbb{R}^m$, which does a stereographic un-projection:
{\small
\begin{equation}
    Q(u) = \frac{1}{||u||}\left[\begin{array}{cccc}
        \frac{1}{2}(||u||^2-1), & u_1, & \ldots, & u_{m-1}
    \end{array}\right]^T
\label{eqn:normalized_unproject}
\end{equation}
}
Note that the un-projection is not actually back to the sphere, but in a way that coordinates 2 through $m$ are a unit vector. Now we can use between $1$ and $n-2$ normalized projections on the representation from the previous case, while still preserving continuity and one-to-one behavior.

For simplicity, we will first demonstrate the case of one stereographic projection. The idea is that we can flatten the representation from Case 3 to a vector and then stereographically project the last $n+1$ components of that vector. Note that we intentionally project as few components as possible, since we found that nonlinearities introduced by the projection can make the learning process more difficult. These nonlinearities are due to the square terms and the division in \eqn{eqn:normalized_unproject}. If $u$ is a vector of length $m$, define the slicing notation $u_{i:j} = (u_i, u_{i+1}, \ldots, u_j)$, and $u_{i:} = u_{i:m}$. Let $M_{(i)}$ be the $i$th column of matrix M. Define a vectorized representation $\gamma(M)$ by dropping the last column of $M$ like in \eqn{eqn:gramschmidt_repr}: $\gamma(M) = [M_{(1)}^T, \ldots, M_{(n-1)}^T]$. Now we can define the mapping to the representation space as:
\begin{equation}
    g_\mathrm{P}(M) = [\gamma_{1:n^2-2n-1}, P(\gamma_{n^2-2n:})]
\label{eqn:project_g}
\end{equation}
Here we have dropped the implicit argument $M$ to $\gamma$ for brevity. Define the mapping to the original space as:
\begin{equation}
    f_\mathrm{P}(u) = f_{\mathrm{GS}}\left([u_{1:n^2-2n-1}, Q(u_{n^2-2n:})]^{(n \times (n-1))}\right)
\label{eqn:project_f}
\end{equation}
Here the superscript $(n \times (n-1))$ indicates that the vector is reshaped to a matrix of the specified size before going through the Gram-Schmidt function $f_{\mathrm{GS}}$. We can now see why \eqn{eqn:normalized_unproject} is normalized the way it is. This is so that projection followed by un-projection can preserve the unit length property of the basis vector that is a column of $M$, and also correctly recover the first component of $Q(\cdot)$, so we get $f_\mathrm{P}(g_\mathrm{P}(M)) = M$ for all $M \in SO(n)$. We can show that $g_\mathrm{P}$ is defined on its domain and continuous by using properties of the orthonormal basis produced by the Gram-Schmidt process $g_{\mathrm{GS}}$. For example, we can show that $\gamma \neq 0$ because components 2 through $n+1$ of $\gamma$ are an orthonormal basis vector, and $N(\gamma)$ will never be equal to the projection point $[1, 0, \ldots, 0, 0]$ that the stereographic projection is made from. It can also be shown that for all $M \in SO(n),$ $f_\mathrm{P}(g_\mathrm{P}(M)) = M$. We show some of these details in the supplemental material.

As a special case, for the 3D rotations, this gives us a 5D representation. This representation is made by using the 6D representation from Case 3, flattening it to a vector, and then using a normalized projection on the last 4 dimensions.

\begin{figure}
    \centering
    \includegraphics[width=3.2in]{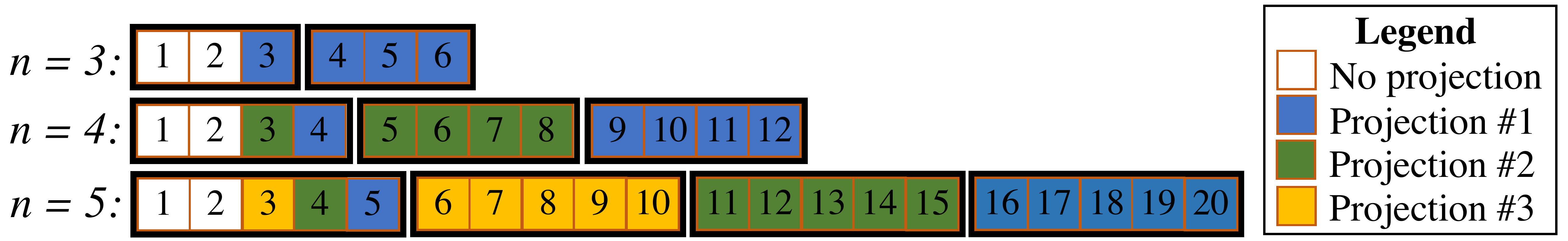}
    \caption{An illustration of how $n-2$ normalized projections can be made to reduce the dimensionality for the representation of $SO(n)$ from Case 3 by $n-2$. In each row we show the dimension $n$, and the elements of the vectorized representation $\gamma(M)$ containing the first $n-1$ columns of $M \in SO(n)$. Each column is length $n$: the columns are grouped by the thick black rectangles. Each unique color specifies a group of inputs for the ``normalized projection" of \eqn{eqn:normalized_project}. The white regions are not projected.}
    
    \label{fig:multi_projection}
    
\end{figure}

We can actually make up to $n-2$ projections in a similar manner, while maintaining continuity of the representation, as follows. As a reminder, the length of $\gamma$, the vectorized result of the Gram-Schmidt process, is $n(n-1)$: it contains $n-1$ basis vectors each of dimension $n$. Thus, we can make $n-2$ projections, where each projection $i=1, \ldots, n-2$ selects the basis vector $i+1$ from $\gamma(M)$, prepends to it an appropriately selected element from the first basis vector of $\gamma(M)$, such as $\gamma_{n+1-i}$, and then projects the result. The resulting projections are then concatenated as a row vector along with the two unprojected entries to form the representation. Thus, after doing $n-2$ projections, we can obtain a continuous representation for $SO(n)$ in $n^2-2n+2$
dimensions. See \fig{fig:multi_projection} for a visualization of the grouping of the elements that can be projected.

\textbf{Other groups: $O(n)$, similarity transforms, quaternions.} In this paper, we focus mainly on the representation of rotations. However, we note that the preceding representations can easily be generalized to $O(n),$ the group of orthogonal $n \times n$ matrices $M$ with $MM^T = M^TM = I$. We can also generalize to the similarity transforms, which we denote as $\text{Sim}(n)$, defined as the affine maps $\rho(x)$ on $\mathbb{R}^n$, $\rho(x) = \alpha R x + u$, where $\alpha>0$, $R$ is an $n \times n$ orthogonal matrix, and $u \in \mathbb{R}^n$~\cite{allen2014motion}. For the orthogonal group $O(n)$, we can use any of the representations in Case 3 or 4, but with an additional component in the representation that indicates whether the determinant is +1 or -1. Then the Gram-Schmidt process in \eqn{eqn:gramschmidt_forward_repr} needs to be modified slightly: if the determinant is -1 then the last vector $b_n$ needs to be negated. Meanwhile, for the similarity transforms, the translation component $u$ can easily be represented as-is. The matrix component $\alpha R$ of the similarity transform can be represented using any of the options in Case 3 or 4. The only needed change is that the Gram-Schmidt process in Equations~\ref{eqn:gramschmidt_forward_repr} or \ref{eqn:project_f} should multiply the final resulting matrix by $\alpha$. The term $\alpha$ is simply the norm of any of the basis vectors input to the Gram-Schmidt process, e.g. $||a_1||$ in \eqn{eqn:gramschmidt_forward_repr}. Clearly, if the projections of Case 4 are used, at least one basis vector must remain not projected, so $\alpha$ can be determined. In the supplemental material, we also explain how one might adapt an existing network that outputs 3D or 4D rotation representations so that it can use our 6D or 5D representations.

\input{Empirical_results}
\vspace{-1.5ex}
\section{Conclusion}
We investigated the use of neural networks to approximate the mappings between various rotation representations. We found empirically that neural networks can better fit continuous representations. For 3D rotations, the commonly used quaternion and Euler angle representations have discontinuities and can cause problems during learning. We present continuous 5D and 6D rotation representations and demonstrate their advantages using an auto-encoder sanity test, as well as real world applications, such as 3D pose estimation and human inverse kinematics.

\section{Acknowledgements}
We thank Noam Aigerman, Kee Yuen Lam, and Sitao Xiang for fruitful discussions; Fangjian Guo, Xinchen Yan, and Haoqi Li for helping with the presentation. This research was conducted at USC and Adobe and was funded by in part by the ONR YIP grant N00014-17-S-FO14, the CONIX Research Center, one of six centers in JUMP, a Semiconductor Research Corporation (SRC) program sponsored by DARPA, the Andrew and Erna Viterbi Early Career Chair, the U.S. Army Research Laboratory (ARL) under contract number W911NF-14-D-0005, Adobe, and Sony. This project was not funded by Pinscreen, nor has it been conducted at Pinscreen or by anyone else affiliated with Pinscreen. The content of the information does not necessarily reflect the position or the policy of the Government, and no official endorsement should be inferred.

{\small
\bibliographystyle{ieee}
\bibliography{egbib}
}


\input{supplemental.tex}

\end{document}

%% file: macros.tex
\newcommand{\ignorethis} [1] {}

\newcommand{\sectnum    } [1] {\ref{#1}}

\newcommand{\sect       } [1] {Section~\sectnum{#1}}

\newcommand{\fignum     } [1] {\ref{#1}}
\newcommand{\fig        } [1] {Figure~\fignum{#1}}
\newcommand{\eqn        } [1] {Equation~(\ref{#1})}

\DeclareMathOperator{\Tr}{Tr}

%% file: abstract.tex
\begin{abstract}
In neural networks, it is often desirable to work with various representations of the same space. For example, 3D rotations can be represented with quaternions or Euler angles. In this paper, we advance a  definition of a continuous representation, which can be helpful for training deep neural networks. We relate this to topological concepts such as homeomorphism and embedding. We then investigate what are continuous and discontinuous representations for 2D, 3D, and n-dimensional rotations. We demonstrate that for 3D rotations, all representations are discontinuous in the real Euclidean spaces of four or fewer dimensions. Thus, widely used representations such as quaternions and Euler angles are discontinuous and difficult for neural networks to learn. We show that the 3D rotations have continuous representations in 5D and 6D, which are more suitable for learning. We also present continuous representations for the general case of the $n$ dimensional rotation group $SO(n)$.
While our main focus is on rotations, we also show that our constructions apply to other groups such as the orthogonal group and similarity transforms. We finally present empirical results, which show that our continuous rotation representations outperform discontinuous ones for several practical problems in graphics and vision, including a simple autoencoder sanity test, a rotation estimator for 3D point clouds, and an inverse kinematics solver for 3D human poses.
\end{abstract}

%% file: introduction.tex
\section{Introduction}

Recently, there has been an increasing number of applications in graphics and vision, where deep neural networks are used to perform regressions on rotations. This has been done for tasks such as pose estimation from images~\cite{do2018deep, xiang2017posecnn} and from point clouds~\cite{gao2018occlusion}, structure from motion~\cite{ummenhofer2017demon}, and skeleton motion synthesis, which generates the rotations of joints in skeletons~\cite{villegas2018neural}. Many of these works represent 3D rotations using 3D or 4D representations such as quaternions, axis-angles, or Euler angles. 

However, for 3D rotations, we found that 3D and 4D representations are not ideal for network regression, when the full rotation space is required. Empirically, the converged networks still produce large errors at certain rotation angles. We believe that this actually points to deeper topological problems related to the continuity in the rotation representations. Informally, all else being equal, discontinuous representations should in many cases be ``harder" to approximate by neural networks than continuous ones. Theoretical results suggest that functions that are smoother~\cite{xu2005simultaneous} or have stronger continuity properties such as in the modulus of continuity~\cite{xu2004essential,chen2013construction} have lower approximation error for a given number of neurons.

Based on this insight, we first present in \sect{sec:continuity_definition} our definition of the continuity of representation in neural networks. We illustrate this definition based on a simple example of 2D rotations. We then connect it to key topological concepts such as homeomorphism and embedding.

Next, we present in \sect{sec:continuity_rotations} a theoretical analysis of the continuity of rotation representations. We first investigate in \sect{sec:continuity_rotations_discontinuous} some discontinuous representations, such as Euler angle and quaternion representations. We show that for 3D rotations, all representations are discontinuous in four or fewer dimensional real Euclidean space with the Euclidean topology. We then investigate in \sect{sec:continuity_rotations_continuous} some continuous rotation representations. For the $n$ dimensional rotation group $SO(n)$, we present a continuous $n^2 - n$ dimensional representation. We additionally present an option to reduce the dimensionality of this representation by an additional $1$ to $n-2$ dimensions in a continuous way. We show that these allow us to represent 3D rotations continuously in 6D and 5D. While we focus on rotations, we show how our continuous representations can also apply to other groups such as orthogonal groups $O(n)$ and similarity transforms.

Finally, in \sect{sec:empirical_results} we test our ideas empirically. We conduct experiments on 3D rotations and show that our 6D and 5D continuous representations always outperform the discontinuous ones for several tasks, including a rotation autoencoder ``sanity test," rotation estimation for 3D point clouds, and 3D human pose inverse kinematics learning. We note that in our rotation autoencoder experiments, discontinuous representations can have up to 6 to 14 times higher mean errors than continuous representations. Furthermore they tend to converge much slower while still producing large errors over 170$^{\circ}$ at certain rotation angles even after convergence, which we believe are due to the discontinuities being harder to fit. 
This phenomenon can also be observed in the experiments on different rotation representations for homeomorphic variational auto-encoding in Falorsi~et~al.~\cite{falorsi2018explorations}, and in practical applications, such as 6D object pose estimation in Xiang~et~al.~\cite{xiang2017posecnn}.

We also show that one can perform direct regression on 3x3 rotation matrices. Empirically this approach introduces larger errors than our 6D representation as shown in \sect{sec:pose_estimation}. Additionally, for some applications such as inverse and forward kinematics, it may be important for the network itself to produce orthogonal matrices. We therefore require an orthogonalization procedure in the network. In particular, if we use a Gram-Schmidt orthogonalization, we then effectively end up with our 6D representation. 

Our contributions are: 1) a definition of continuity for rotation representations, which is suitable for neural networks; 2) an analysis of discontinuous and continuous representations for 2D, 3D, and $n$-D rotations; 3) new formulas for continuous representations of $SO(3)$ and $SO(n)$; 4) empirical results supporting our theoretical views and that our continuous representations are more suitable for learning.

%% file: related_work.tex
\section{Related Work}

In this section, we will first establish some context for our work in terms of neural network approximation theory. Next, we discuss related works that investigate the continuity properties of different rotation representations. Finally, we will report the types of rotation representations used in previous learning tasks and their performance.

\textbf{Neural network approximation theory.} We review a brief sampling of results from neural network approximation theory. Hornik~\cite{hornik1991approximation} showed that neural networks can approximate functions in the $L^p$ space to arbitrary accuracy if the $L^p$ norm is used. Barron~et~al.~\cite{barron1993universal} showed that if a function has certain properties in its Fourier transform, then at most $O(\epsilon^{-2})$ neurons are needed to obtain an order of approximation $\epsilon$. Chapter 6.4.1 of LeCun~et~al.~\cite{lecun2015deep} provides a more thorough overview of such results. We note that results for continuous functions indicate that functions that have better smoothness properties can have lower approximation error for a particular number of neurons~\cite{xu2004essential,chen2013construction,xu2005simultaneous}. For discontinuous functions, Llanas et al.~\cite{llanas2008constructive} showed that a real and piecewise continuous function can be approximated in an almost uniform way. However, Llanas et al.~\cite{llanas2008constructive} also noted that piecewise continuous functions when trained with gradient descent methods require many neurons and training iterations, and yet do not give very good results. These results suggest that continuous rotation representations might perform better in practice. 



%

\textbf{Continuity for rotations.} Grassia~et~al.~\cite{grassia1998practical} pointed out that Euler angles and quaternions are not suitable for orientation differentiation and integration operations and proposed exponential map as a more robust rotation representation. Saxena~et~al.~\cite{saxena2009learning} observed that the Euler angles and quaternions cause learning problems due to discontinuities. However, they did not propose general rotation representations other than direct regression of 3x3 matrices, since they focus on learning representations for objects with specific symmetries. 

\textbf{Neural networks for 3D shape pose estimation.}
Deep networks have been applied to estimate the 6D poses of object instances from RGB images, depth maps or scanned point clouds. Instead of directly predicting 3x3 matrices that may not correspond to valid rotations, they typically use more compact rotation representations such as quaternion~\cite{xiang2017posecnn,kendall2015posenet,kendall2017geometric} or axis-angle~\cite{ummenhofer2017demon,gao2018occlusion,do2018deep}. In PoseCNN~\cite{xiang2017posecnn}, the authors reported a high percentage of errors between 90$^\circ$ and 180$^\circ$, and suggested that this is mainly caused by the rotation ambiguity for some symmetric shapes in the test set. However, as illustrated in their paper, the proportion of errors between 90$^\circ$ to 180$^\circ$ is still high even for non-symmetric shapes. In this paper, we argue that discontinuity in these representations could be one cause of such errors.

\textbf{Neural networks for inverse kinematics.}
Recently, researchers have been interested in training neural networks to solve inverse kinematics equations. This is because such networks are faster than traditional methods and differentiable so that they can be used in more complex learning tasks such as motion re-targeting~\cite{villegas2018neural} and video-based human pose estimation~\cite{kanazawa2018end}. Most of these works represented rotations using quaternions or axis-angle~\cite{hsu2018quatnet,kanazawa2018end}. Some works also used other 3D representations such as Euler angles and Lie algebra~\cite{kanazawa2018end, zhou2016deep}, and penalized the joint position errors. 
Csiszar~et~al.~\cite{csiszar2017solving} designed  networks to output the sine and cosine of the Euler angles for solving the inverse kinematics problems in robotic control. Euler angle representations are discontinuous for $SO(3)$ and can result in large regression errors as shown in the empirical test in \sect{sec:empirical_results}. However, those authors limited the rotation angles to be within a certain range, which avoided the discontinuity points and thus achieved very low joint alignment errors in their test. However, many real-world tasks require the networks to be able to output the full range of rotations. In such cases,  continuous rotation representations will be a better choice.

%% file: Empirical_results.tex
\section{Empirical Results}
\begin{figure*}
    \centering
    \includegraphics[width=0.96\textwidth]{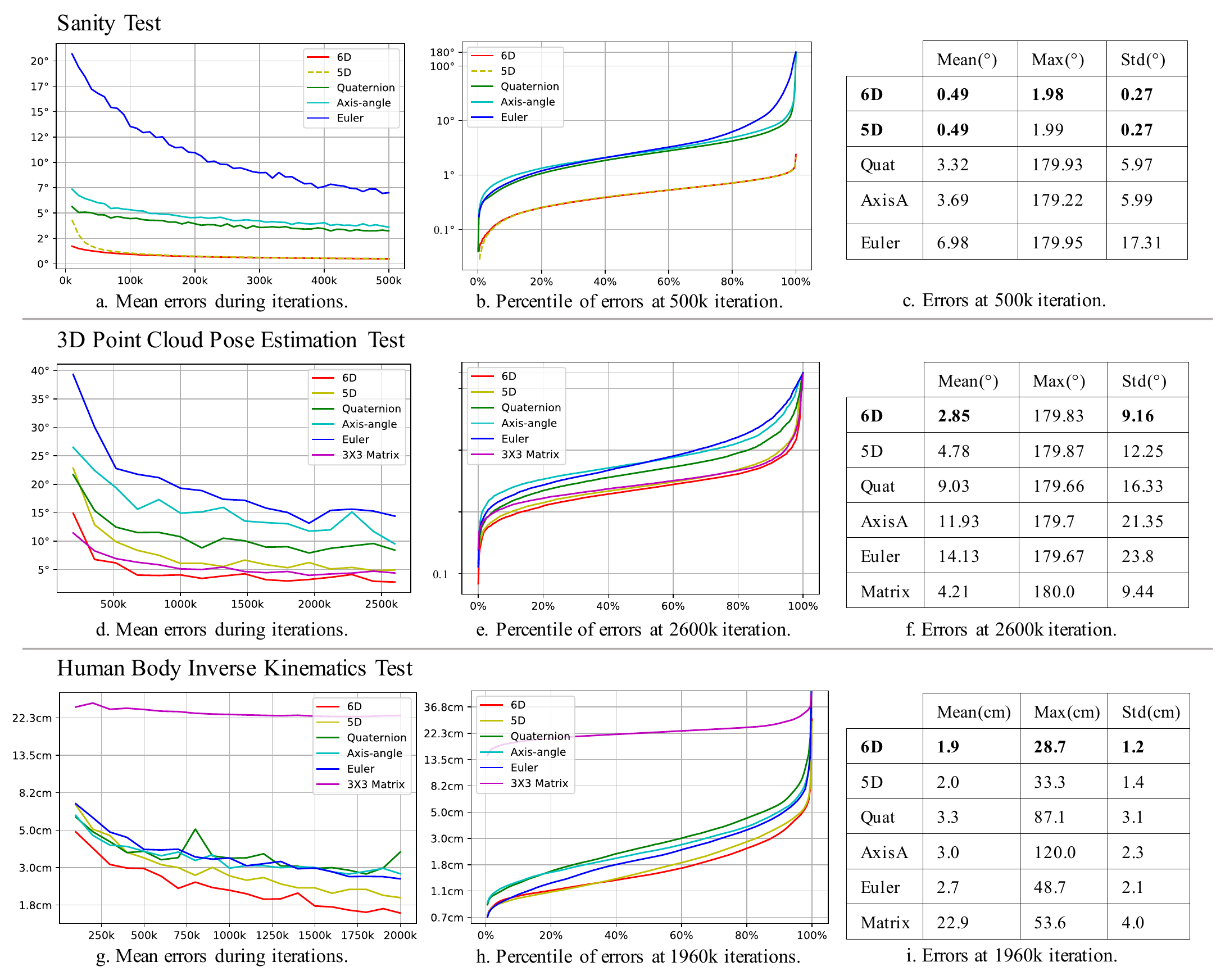}
    \vspace{-0.1cm}
    \caption{Empirical results. In (b), (e), (h) we plot on the x axis a percentile $p$ and on the y axis the error at the given percentile $p$.}
    \vspace{-0.5cm}
    \label{fig:empirical_result}
\end{figure*}

\label{sec:empirical_results}
We investigated different rotation representations and found that those with better continuity properties work better for learning. We first performed a sanity test and then experimented on two real world applications to show how continuity properties of rotation representations influence the learning process.
\subsection{Sanity Test}
\label{sec:sanity_test}
We first perform a sanity test using an auto-encoder structure.
We use a multi-layer perceptron (MLP) network as an encoder to map $SO(3)$ to the chosen representation $R$. We test our proposed 6D and 5D representations, quaternions, axis-angle, and Euler angles. The encoder network contains four fully-connected layers, where hidden layers have 128 neurons and Leaky ReLU activations. The fixed ``decoder" mapping $f: R \mapsto SO(3)$ is defined in Section~\ref{sec:continuity_rotations}.

For training, we compute the loss using the L2 distance between the input $SO(3)$ matrix $M$ and the output $SO(3)$ matrix $M'$: note that this is invariant to the particular representation used, such as quaternions, axis-angle, etc. We use Adam optimization with batch size 64 and learning rate $10^{-5}$ for the first $10^4$ iterations and $10^{-6}$ for the remaining iterations. For sampling the input rotation matrices during training, we uniformly sample axes and angles. We test the networks using $10^5$ rotation matrices generated by randomly sampling axes and angles and calculate geodesic errors between the input and the output rotation matrices. The geodesic error is defined as the minimal angular difference between two rotations, written as 
{\small
\begin{equation}
L_{\mathrm{angle}} = \text{cos}^{-1}((M''_{00}+M''_{11}+M''_{22}-1)/2)
\end{equation}
\begin{equation}
M''= MM'^{-1}
\end{equation}
}
Figure~\ref{fig:empirical_result}(a) illustrates the mean geodesic errors for different representations as training progresses. Figure~\ref{fig:empirical_result}(b) illustrates the percentiles of the errors at 500k iterations. The results show that the 6D and 5D representations have similar performance with each other. They converge much faster than the other representations and produce smallest mean, maximum and standard deviation of errors. The Euler angle representation performs the worst, as shown in Table (c) in Figure~\ref{fig:empirical_result}. For the quaternion, axis angle and Euler angle representations, the majority of the errors fall under 25$^\circ$, but certain test samples still produce errors up to 180$^\circ$. The proposed 6D and 5D representations do not produce errors higher than 2$^\circ$. We conclude that using continuous rotation representations for network training leads to lower errors and faster convergence.

In Appendix \ref{sec:additional_sanity}, we report additional results, where we trained using a geodesic loss, uniformly sampled $SO(3)$, and compared to 3D Rodriguez vectors and quaternions constrained to one hemisphere~\cite{kendall2017geometric}. Again, our continuous representations outperform common discontinuous ones.

\vspace{-5pt}
\subsection{Pose Estimation for 3D Point Clouds}
\label{sec:pose_estimation}
\vspace{-5pt}

In this experiment, we test different rotation representations on the task of estimating the rotation of a target point cloud from a reference point cloud. The inputs of the networks are the reference and target point clouds $P_\mathrm{r}, P_\mathrm{t}\in \mathbb{R}^{N\times3}$, where $N$ is the number of points. The network output is the estimated rotation $R \in \mathbb{R}^{D}$ between $P_\mathrm{r}$ and $P_\mathrm{t}$, where $D$ is the dimension of the chosen representation.

We employ a weight-sharing Siamese network where each half is a simplified PointNet structure~\cite{qi2017pointnet}, $\Phi : \mathbb{R}^{N\times3} \mapsto \mathbb{R}^{1024}$. The simplified PointNet uses a 4-layer MLP of size $3\times64\times128\times1024$ to extract features for each point and then applies max pooling across all points to produce a single feature vector $z$. One half of the Siamese network maps the reference point cloud to a feature vector $z_\mathrm{r} = \Phi(P_\mathrm{r})$ and the other half maps the target point cloud to $z_\mathrm{t} = \Phi(P_\mathrm{t})$. Then we concatenate $z_\mathrm{r}$ and $z_\mathrm{t}$ and pass this through another MLP of size $2048\times512\times512\times D$ to produce the $D$ dimensional rotation representation. Finally, we transform the rotation representations to $SO(3)$ with one of the mapping functions $f$ defined in Section~\ref{sec:continuity_rotations}.

We train the network with 2,290 airplane point clouds from ShapeNet~\cite{chang2015shapenet}, and test it with 400 held-out point clouds augmented with 100 random rotations. At each training iteration, we randomly select a reference point cloud and transform it with 10 randomly-sampled rotation matrices to get 10 target point clouds. We feed the paired reference-target point clouds into the Siamese network and minimize the L2 loss between the output and the ground-truth rotation matrices.

We trained the network with $2.6\times10^6$ iterations. Plot (d) in Figure~\ref{fig:empirical_result} shows the mean geodesic errors as training progresses. Plot (e) and Table (f) show the percentile, mean, max and standard deviation of errors. Again, the 6D representation has the lowest mean and standard deviation of errors with around 95\% of errors lower than 5$^\circ$, while Euler representation is the worst with around 10\% of errors higher than 25$^\circ$. Unlike the sanity test, the 5D representation here performs worse than the 6D representation, but outperforms the 3D and 4D representations. We hypothesize that the distortion in the gradients caused by the stereographic projection makes it harder for the network to do the regression. 
Since the ground-truth rotation matrices are available, we can directly regress the $3\times3$ matrix using L2 loss. During testing, we use the Gram-Schmidt process to transform the predicted matrix into $SO(3)$ and then report the geodesic error (see the bottom row of Table (f) in Figure~\ref{fig:empirical_result}). We hypothesize that the reason for the worse performance of the $3\times3$ matrix compared to the 6D representation is due to the orthogonalization post-process, which introduces errors.

\subsection{Inverse Kinematics for Human Poses}  
In this experiment, we train a neural network to solve human pose inverse kinematics (IK) problems. Similar to the method of Villegas~et~al.~\cite{villegas2018neural} and Hsu~et~al.~\cite{hsu2018quatnet}, our network takes the joint positions of the current pose as inputs and predicts the rotations from the T-pose to the current pose. We use a fixed forward kinematic function to transform predicted rotations back to joint positions and penalize their L2 distance from the ground truth. Previous work for this task used quaternions. We instead test on different rotation representations and compare their performance. 

The input contains the 3D positions of the $N$ joints on the skeleton marked as $P=(p_1, p_2, p_3,...,p_N)$, $p_i=(x,y,z)^\intercal$. The output of the network are the rotations of the joints in the chosen representation $R=(r_1, r_2, r_3,...,r_N), r_i\in \mathbb{R}^D$, where $D$ is the dimension of the representation.

We train a four-layer MLP network that has 1024 neurons in hidden layers with the L2 reconstruction loss $L=||P-P'||^2_2$, where $P'=\Pi(T, R)$. Here $\Pi$ is the forward kinematics function which takes as inputs the ``T" pose of the skeleton and the predicted joints rotations, and outputs the 3D positions of the joints. Due to the recursive computational structure of forward kinematics, the accuracy of the hip orientation is critical for the overall skeleton pose prediction and thus the joints adjacent to the hip contribute more weight to the loss (10 times higher than other joints).

We use the CMU Motion Capture Database~\cite{CMUdataset} for training and testing because it contains complex motions like dancing and martial arts, which cover a wide range of joint rotations. We picked in total 865 motion clips from 37 motion categories. We randomly chose 73 clips for testing and the rest for training. We fix the global position of the hip so that we do not need to worry about predicting the global translation. The whole training set contains $1.14 \times 10^6$ frames of human poses and the test set contains $1.07 \times 10^5$ frames of human poses. We train the networks with 1,960k iterations with batch size 64. During training, we augmented the poses with random rotation along the y-axis. We augmented each instance in the test set with three random rotations along the y-axis as well. The results, as displayed in subplots (g), (h)  and (i) in Figure~\ref{fig:empirical_result}, show that the 6D representation performs the best with the lowest errors and fastest convergence. The 5D representation has similar performance as the 6D one. On the contrary, the 4D and 3D representations have higher average errors and higher percentages of big errors that exceed 10 cm. 

We also perform the test of using the 3$\times$3 matrix without orthogonalization during training and using the Gram-Schmidt process to transform the predicted matrix into SO(3) during testing. We find this method creates huge errors as reported in the bottom line of Table (i) in Figure \ref{fig:empirical_result}. One possible reason for this bad performance is that the 3$\times$3 matrix may cause the bone lengths to scale during the forward kinematics process.
In Appendix \ref{sec:visual_IK}, we additionally visualize some human body poses for quaternions and our 6D representation.

%% file: supplemental.tex
\appendix

\section{Overview of the Supplemental Document}

In this supplemental material, in \sect{sec:sixd_repr}, we first give explicitly our 6D representation for the 3D rotations. We then prove formally in \sect{sec:proof} that the formula of the 5D representation as defined in Case 4 of \sect{sec:continuity_rotations_continuous} satisfies all of the properties of a continuous representation. 
Next, we discuss quaternions in more depth. We present in \sect{sec:unit_quat} a result that we elided from the main paper due to space limitations: that the unit quaternions are also a discontinuous representation for the 3D rotations. Then, in \sect{sec:quat_rep}, we show how the continuous 5D and 6D representations can interact with common discontinuous angle representations such as quaternions. Next, we visualize in \sect{sec:vis} discontinuities that are present in some of the representations. We finally present in \sect{sec:additional_empirical} some additional empirical results.

\section{6D Representation for the 3D Rotations}
\label{sec:sixd_repr}
The mapping from SO(3) to our 6D representation is:
\begin{equation}
g_{GS}\left(\left[ {\begin{array}{ccc}
   \vertbar & \vertbar & \vertbar \\
   a_1      & a_2      & a_3 \\
   \vertbar &  \vertbar & \vertbar
  \end{array} } \right]\right) = 
  \left[ {\begin{array}{ccc}
   \vertbar & \vertbar \\
   a_1      & a_{2} \\
   \vertbar & \vertbar
  \end{array} } \right]
\end{equation}

\vspace{-0.2cm}

The mapping from our 6D representation to SO(3) is:
\begin{equation}
f_{GS}\left(\left[ {\begin{array}{ccc}
   \vertbar & \vertbar \\
   a_1      & a_2 \\
   \vertbar & \vertbar
  \end{array} } \right]\right) = 
  \left[ {\begin{array}{ccc}
   \vertbar & \vertbar & \vertbar \\
   b_1 & b_2 & b_3 \\
   \vertbar & \vertbar & \vertbar
  \end{array} } \right]
\end{equation}
\begin{equation}
b_i = \left[ \begin{cases}
N(a_1) & \text{if } i=1 \\
N(a_2 - (b_{1} \boldsymbol{\cdot} a_2) b_1) & \text{if } i=2 \\
b_1 \times b_2 & \text{if } i = 3
\end{cases}
\right]^T
\end{equation}

\section{Proof that Case 4 gives a Continuous Representation}
\label{sec:proof}

Here we show that the functions $f_P, g_P$ presented in Case 4 of \sect{sec:continuity_rotations_continuous} are a continuous representation. We now prove some properties needed to show a continuous representation: that  $g_P$ is defined on its domain and continuous, and that $f_P(g_P(M)) = M$ for all $M \in SO(n)$. In these proofs, we use $0$ to denote the zero vector in the appropriate Euclidean space. We also use the same slicing notation from the main paper. That is, if $u$ is a vector of length $m$, define the slicing notation $u_{i:j} = (u_i, u_{i+1}, \ldots, u_j)$, and $u_{i:} = u_{i:m}$.

\textbf{Proof that $g_P$ is defined on $SO(n)$.}
Suppose $M \in SO(n)$. The same as we did for \eqn{eqn:project_g} in the main paper, define a vectorized representation $\gamma(M)$ by dropping the last column of $M$: $\gamma(M) = [M_{(1)}^T, \ldots, M_{(n-1)}^T]$, where $M_{(i)}$ indicates the $i$th column of $M$. Following \eqn{eqn:normalized_project}, which defines the \emph{normalized projection}, and \eqn{eqn:project_g}, let $v = \gamma_{n^2-2n:}/||\gamma_{n^2-2n:}||$. The only way that $g_P(M)$ could not be defined is if the normalized projection $P(v)$ is not defined, which requires $v_1=1$. However, if $v_1 = 1$, then because $v$ is unit length, it follows that $\gamma_{n^2-2n+1:}$ has length zero. But $\gamma_{n^2-2n+1:}$ is a column vector from $M \in SO(n),$ and therefore has unit length. We conclude that $v_1 \neq 1$ and $g_P$ is defined on $SO(n)$.

\textbf{Proof that $g_P$ is continuous.} This case is trivial because $g_P$ is the composition of functions that are continuous on their domains and thus is also continuous on its domain.

\textbf{Lemma 1.} We claim that if $u \in \mathbb{R}^m$ and $||u_{2:}|| = 1$, then $Q(P(u)) = u$. We now prove this. We have $||u|| = \sqrt{1+u_1^2}$. Then we find by \eqn{eqn:normalized_project} that
\begin{equation}
||P(u)|| = \frac{1}{||u||-u_1} = \frac{1}{\sqrt{1+u_1^2}-u_1}
\end{equation}

Now let $b = Q(P(u)).$ Components 2 through $m$ of $b$ are $P(u)/||P(u)||,$ but this is just $u_{2:}$. Next, consider $b_1$, the first component of $b$:
\begin{align}
    b_1 &= \frac{1}{2}\left[||P(u)||-\frac{1}{||P(u)||}\right] \\
    &= \frac{1}{2}\left[\frac{1-(1+u_1^2)+2\sqrt{1+u_1^2}u_1-u_1^2}{\sqrt{1+u_1^2} - u_1}\right] \\
    &= u_1
\end{align}

We find that $b = u,$ so $Q(P(u)) = u$.

\textbf{Proof that $f_P(g_P(M)) = M$ for all $M \in SO(n)$.}
For the term $f_{GS}(A)$ of \eqn{eqn:project_f}, this is defined on $\mathbb{R}^{n(n-1)}\setminus D$. Here $A$ is the matrix argument to $f_{GS}$ in \eqn{eqn:project_f}, and $D$ is the set where the dimension of the span of $A$ is less than $n-1$.  Let $M \in SO(n)$. The same as before, let $\gamma(M)$ be the vectorized representation of $M$, which drops the last column. By Lemma 1, $Q(P(\gamma_{n^2-2n:})) = \gamma_{n^2-2n:}$. Thus by \eqn{eqn:project_f}, we have $f_P(g_P(M)) = f_{GS}(\gamma^{(n \times n-1)}) = f_{GS}(g_{GS}(M)) = M$.  

\section{The Unit Quaternions are a Discontinuous Representation for the 3D Rotations}
\label{sec:unit_quat}

In Case 2 of \sect{sec:continuity_rotations}, we showed that the quaternions are not a continuous representation for the 3D rotations. We intentionally used a simpler formulation for the quaternions, which is easier to understand and also saves space in the paper, due to its quaternions in general not being unit length. However, an attentive reader might wonder what happens if we use the unit quaternions: is the discontinuity removable? However, we show here that the unit quaternions are also not a continuous representation for the 3D rotations.

We use a mapping $g_u$ to map $SO(3)$ to the unit quaternions, which we consider as the Euclidean space $\mathbb{R}^4$. We use the formula by \cite{WikiQuaternion,baker2018}:
\begin{equation}
g_u(M) = \left[ {\begin{array}{c}
   \scriptstyle{\mathrm{copysign}(\frac{1}{2}\sqrt{1+M_{11}-M_{22}-M_{33}}, M_{32}-M_{23})} \\
   \scriptstyle{\mathrm{copysign}(\frac{1}{2}\sqrt{1-M_{11}+M_{22}-M_{33}}, M_{13}-M_{31})} \\
   \scriptstyle{\mathrm{copysign}(\frac{1}{2}\sqrt{1-M_{11}-M_{22}+M_{33}}, M_{21}-M_{12})} \\
   \scriptstyle{\frac{1}{2}\sqrt{1+M_{11}+M_{22}+M_{33}}}
  \end{array} } \right]
\end{equation}

Here $\mathrm{copysign}(a, b) = \mathrm{sgn}(b)|a|$. Now consider the following matrix in $SO(3),$ which is parameterized by $\theta$:

\begin{equation}
B(\theta) = \left[ {\begin{array}{ccc}
   \cos(\theta) & -\sin(\theta) & 0 \\
   \sin(\theta) & \cos(\theta) & 0 \\
   0 & 0 & 1
  \end{array} } \right].
\end{equation}

By substitution, we can find the components of $g_u(B(\theta))$ as a function of $\theta$. For example, as $\theta \to \pi-$, the third component is $\sqrt{(1-\cos(\theta))/2} = 1$. Meanwhile, as $\theta \to \pi+$, the third component is $-\sqrt{(1-\cos(\theta))/2} = -1$. We conclude that the unit quaternions are not a continuous representation.

A similar representation is the Cayley transformation~\cite{WikiCayley} which has a different scaling to the unit quaternion, where $w=1$ and the vector $(x,y,z)$ is the unit axis of rotation scaled by $tan(\theta/2)$. The limit goes to infinity when approaching $180^\circ$. Thus, it is not a  representation for $SO(3)$.

\section{Interaction Between 5D and 6D Continuous Representations and Discontinuous Ones}
\label{sec:quat_rep}

In some cases, it may be convenient to use a common 3D or 4D angle representation, such as the quaternions or Euler angles. For example, the quaternions may be useful when interpolating between two rotations in $SO(3)$, or when there is an existing neural network that already accepts quaternion inputs. However, as we showed in the main paper Case 2 of \sect{sec:continuity_rotations_discontinuous}, all 3D and 4D representations for rotations are discontinuous.

One solution for the above conundrum is to simply convert as needed from the continuous 5D and 6D representations that we presented in Case 3 and 4 of \sect{sec:continuity_rotations_continuous} to the desired representation. For concreteness, suppose the desired representation is the quaternions. Assume that any conversions done in the network are only in the direction that maps to the quaternions. Then the associated mapping in the opposite direction (i.e. from quaternions to the 5D or 6D representation) is continuous. If losses are applied only at points in the network where the representation is continuous (e.g. on the 5D or 6D representations), then the learning should not suffer from discontinuity problems. One can convert from the 5D or 6D representation to quaternions by first applying \eqn{eqn:gramschmidt_repr} or \eqn{eqn:project_g} and then using Equation (4). Of course, one could also make a similar argument for other discontinuous but popular angle representations such as Euler angles.

\section{Visualizing Discontinuities in 3D Rotations}
\label{sec:vis}

\begin{figure}
    \centering
    \begin{tabular}{ccc}
        \multicolumn{3}{c}{$SO(3)$} \vspace{0.3cm} \\
        \includegraphics[width=0.93in, trim=0.4in 0.4in 0.4in 0.8in]{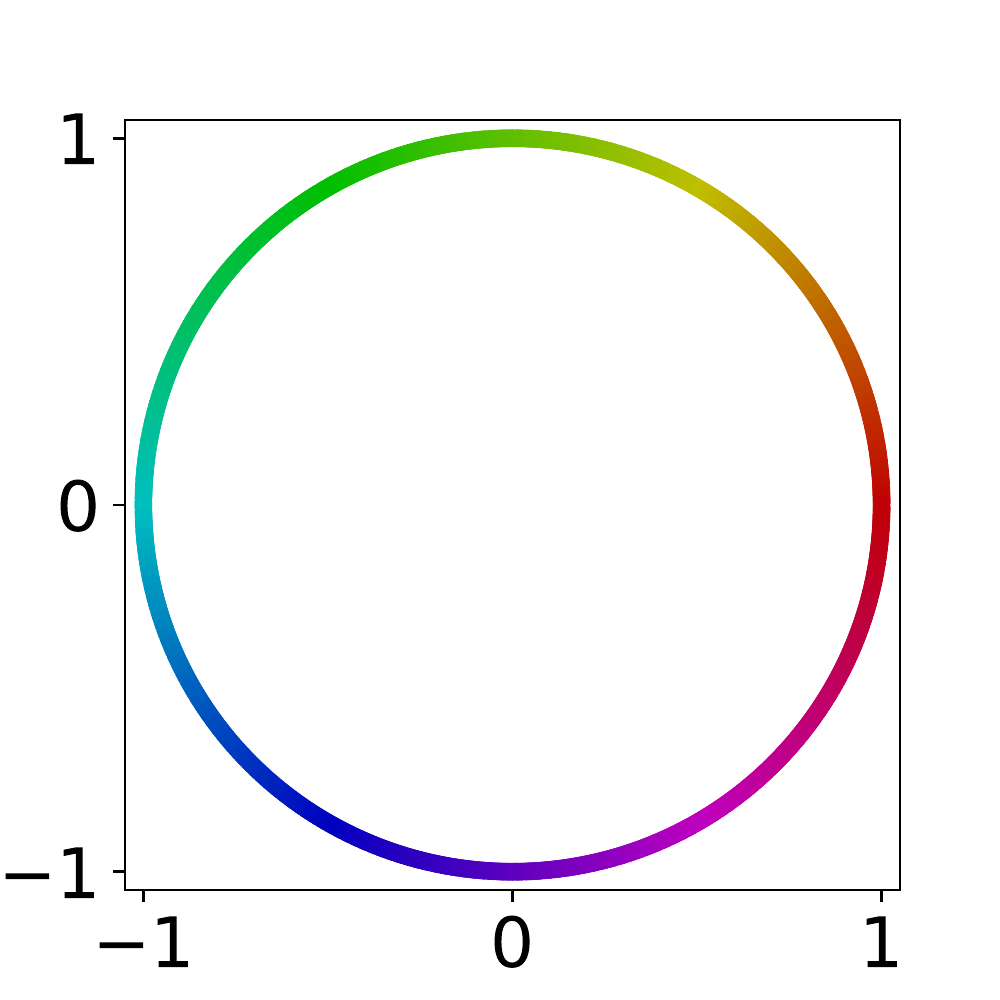} &
        \includegraphics[width=0.93in, trim=0.4in 0.4in 0.4in 0.8in]{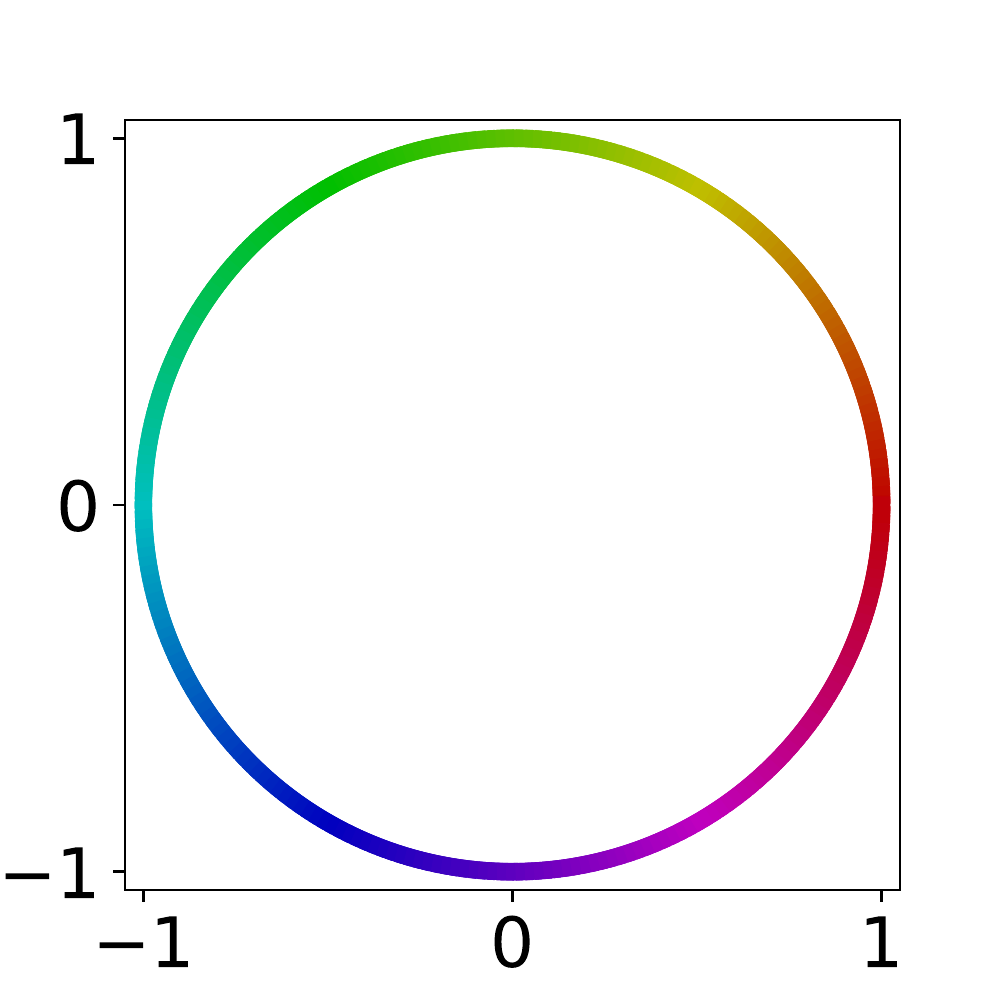} &
        \includegraphics[width=0.93in, trim=0.4in 0.4in 0.4in 0.8in]{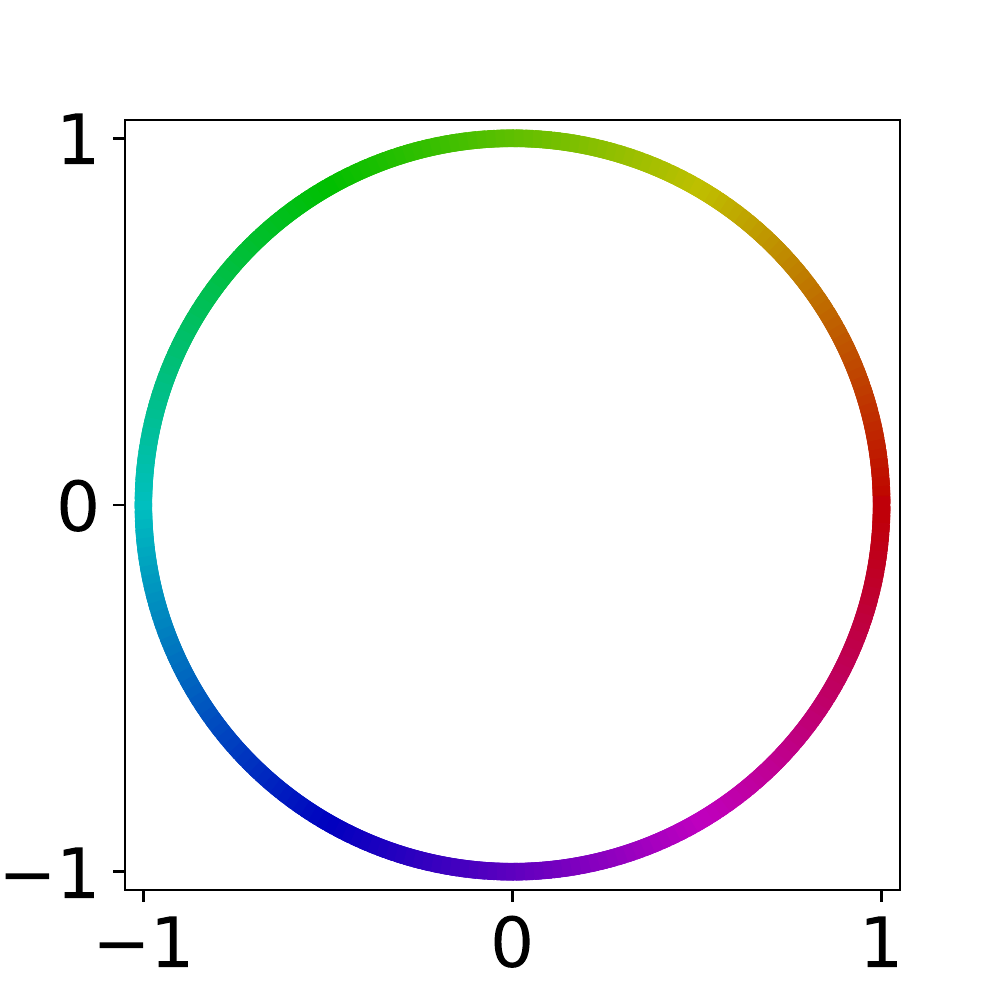} \vspace{0.2cm} \\
        \multicolumn{3}{c}{6D} \vspace{0.3cm} \\
        \includegraphics[width=0.93in, trim=0.4in 0.4in 0.4in 0.8in]{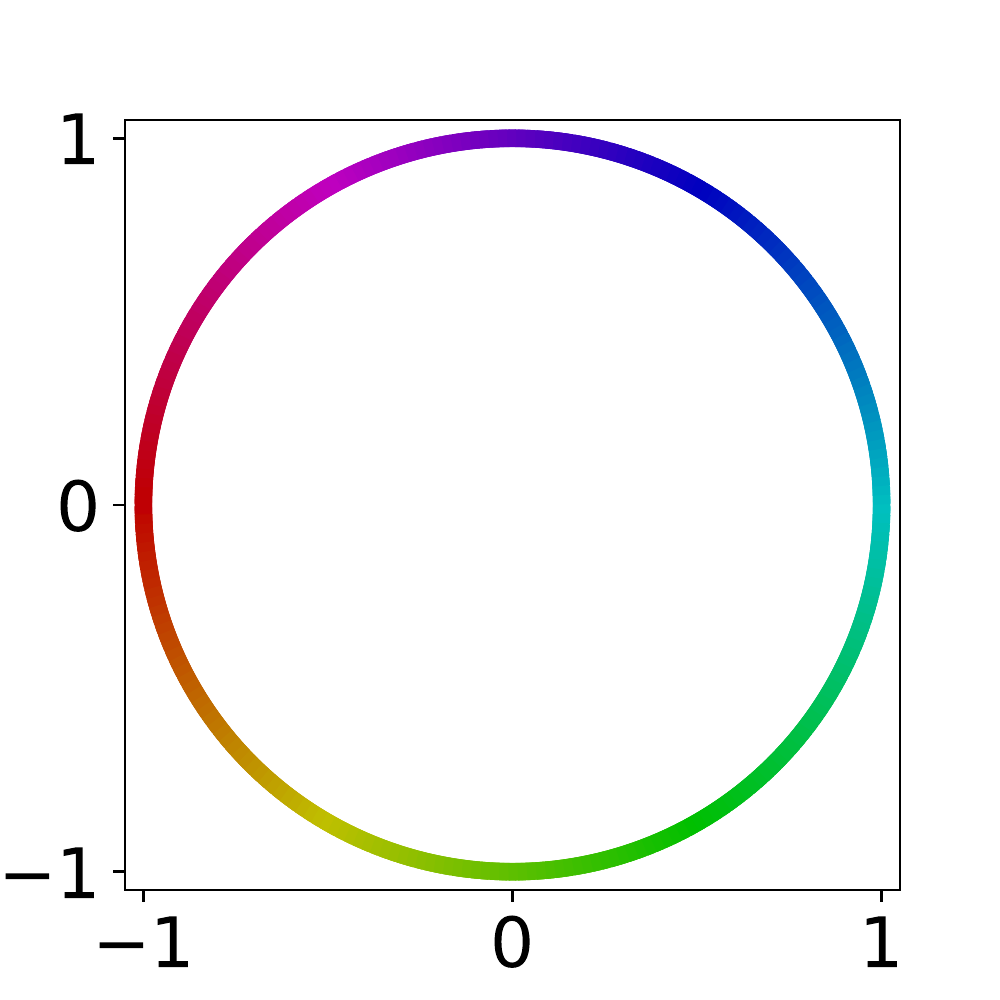} &
        \includegraphics[width=0.93in, trim=0.4in 0.4in 0.4in 0.8in]{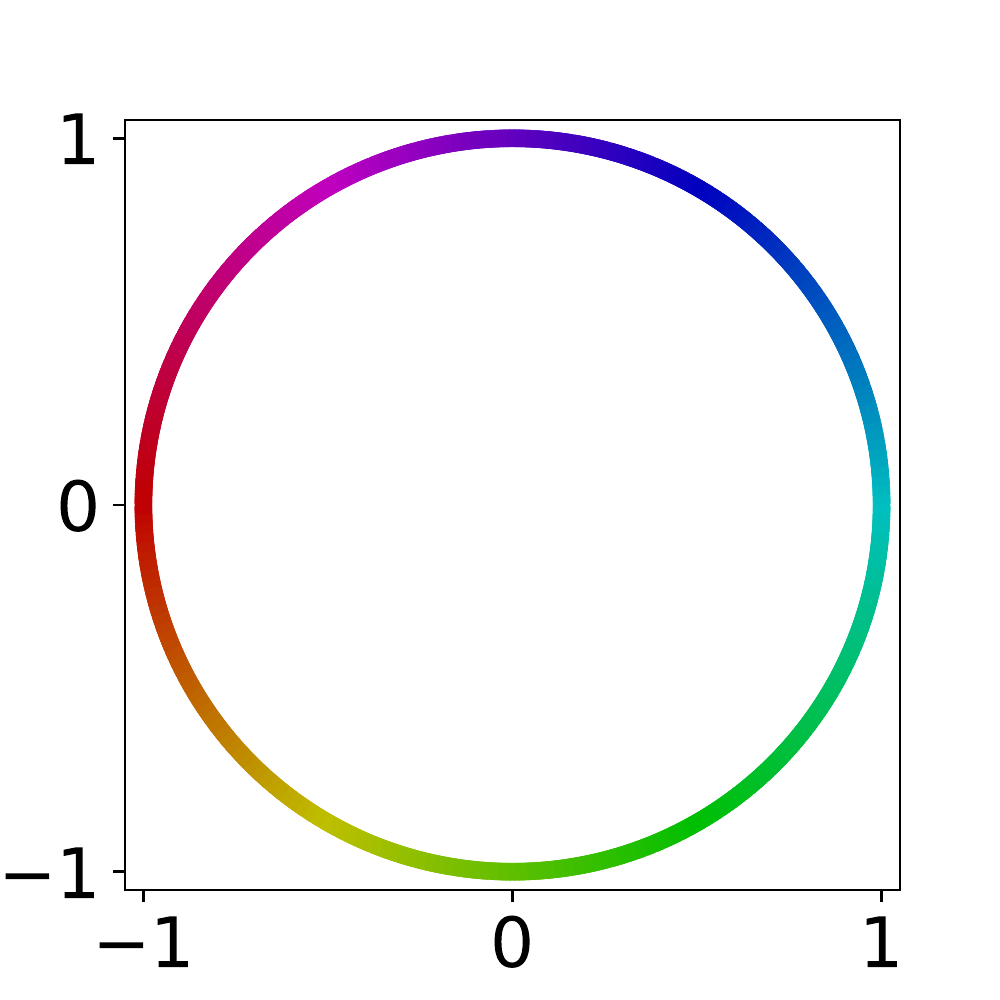} &
        \includegraphics[width=0.93in, trim=0.4in 0.4in 0.4in 0.8in]{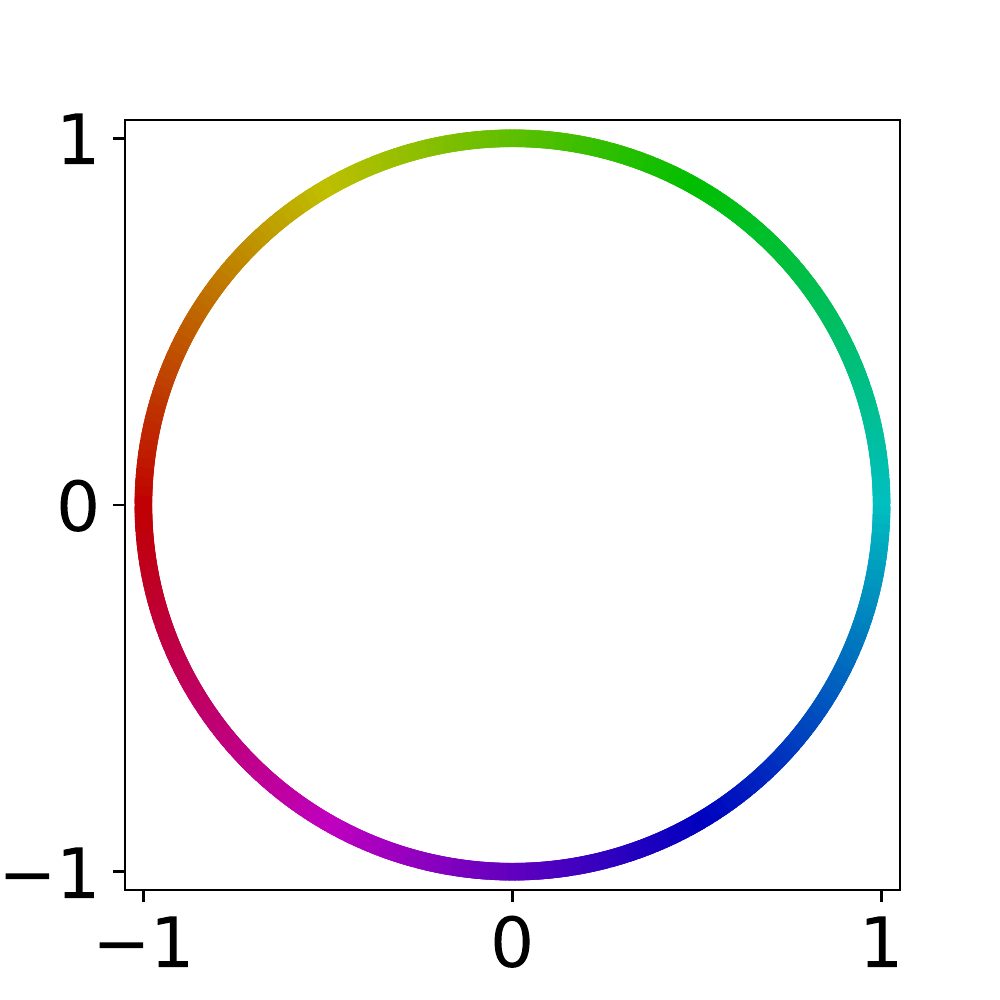} \vspace{0.2cm} \\
        \multicolumn{3}{c}{5D} \vspace{0.3cm} \\
        \includegraphics[width=0.93in, trim=0.4in 0.4in 0.4in 0.8in]{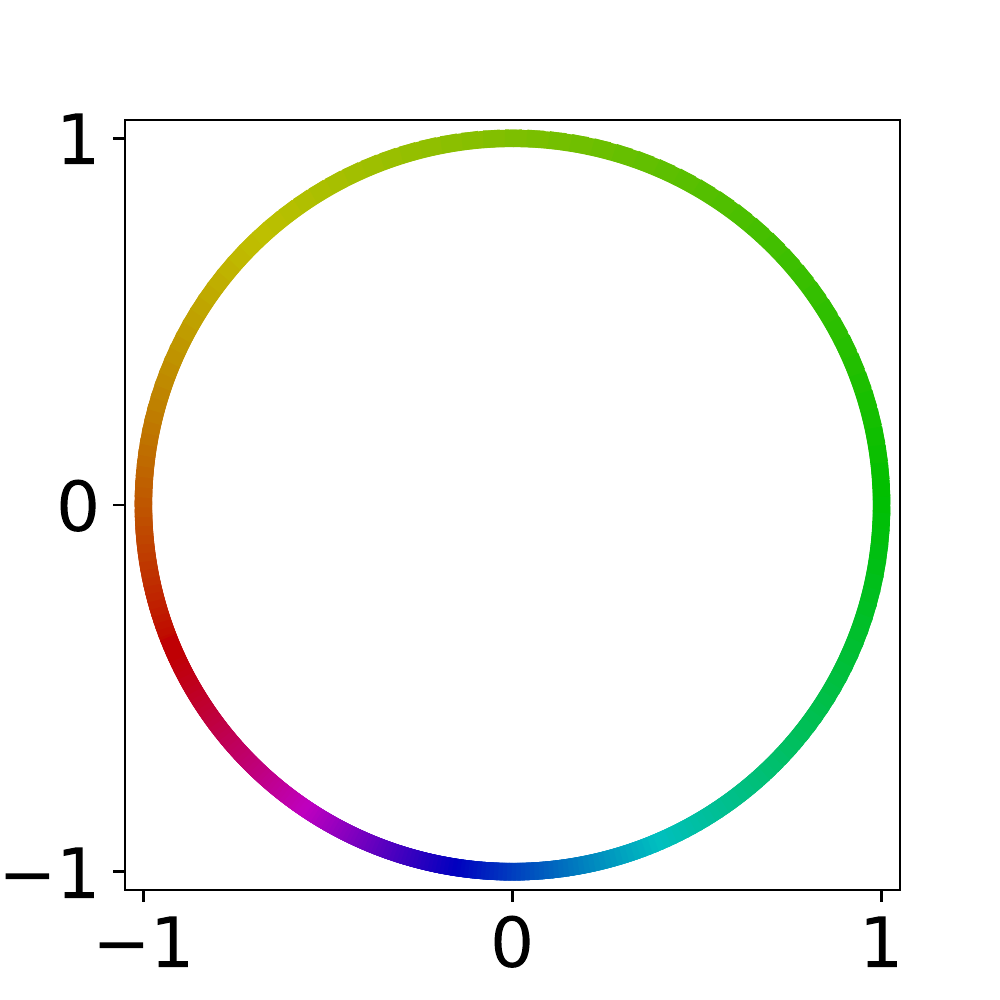} &
        \includegraphics[width=0.93in, trim=0.4in 0.4in 0.4in 0.8in]{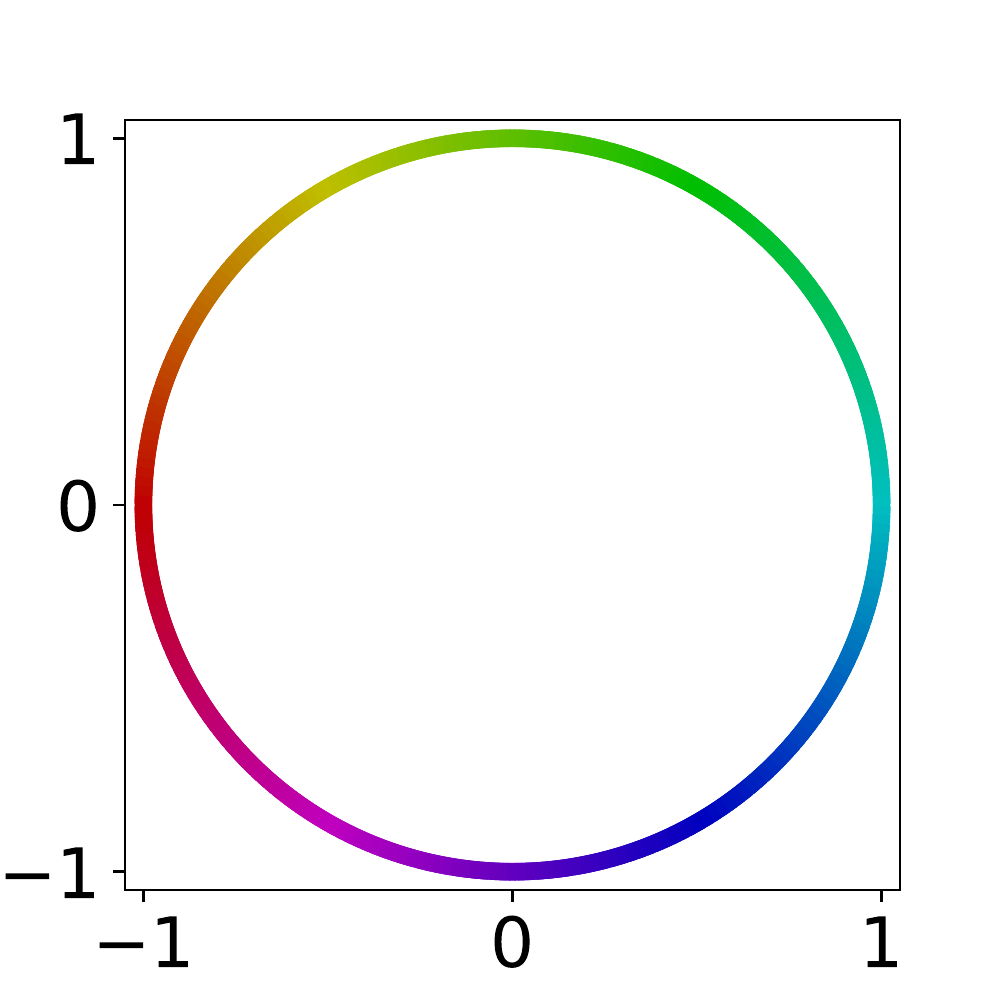} &
        \includegraphics[width=0.93in, trim=0.4in 0.4in 0.4in 0.8in]{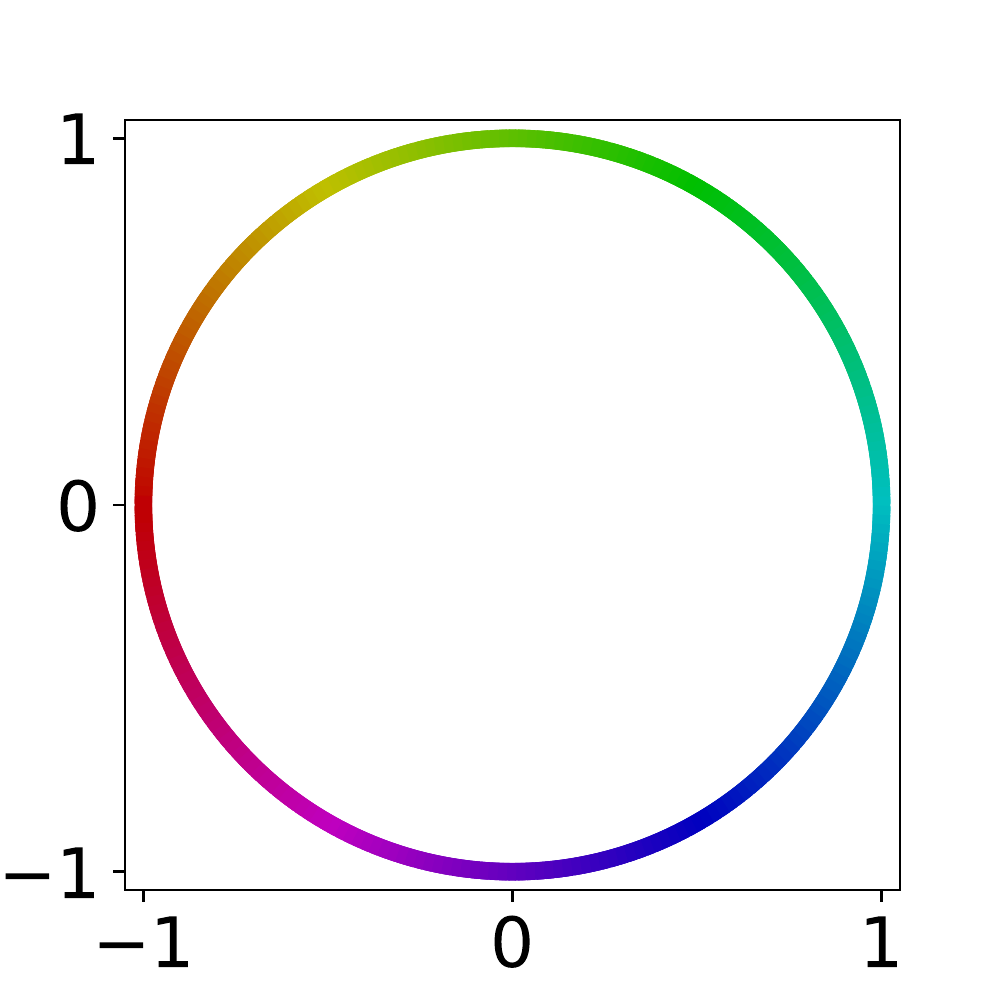} \vspace{0.2cm} \\
        \multicolumn{3}{c}{Unit quaternions} \vspace{0.3cm} \\
        \includegraphics[width=0.93in, trim=0.4in 0.4in 0.4in 0.8in]{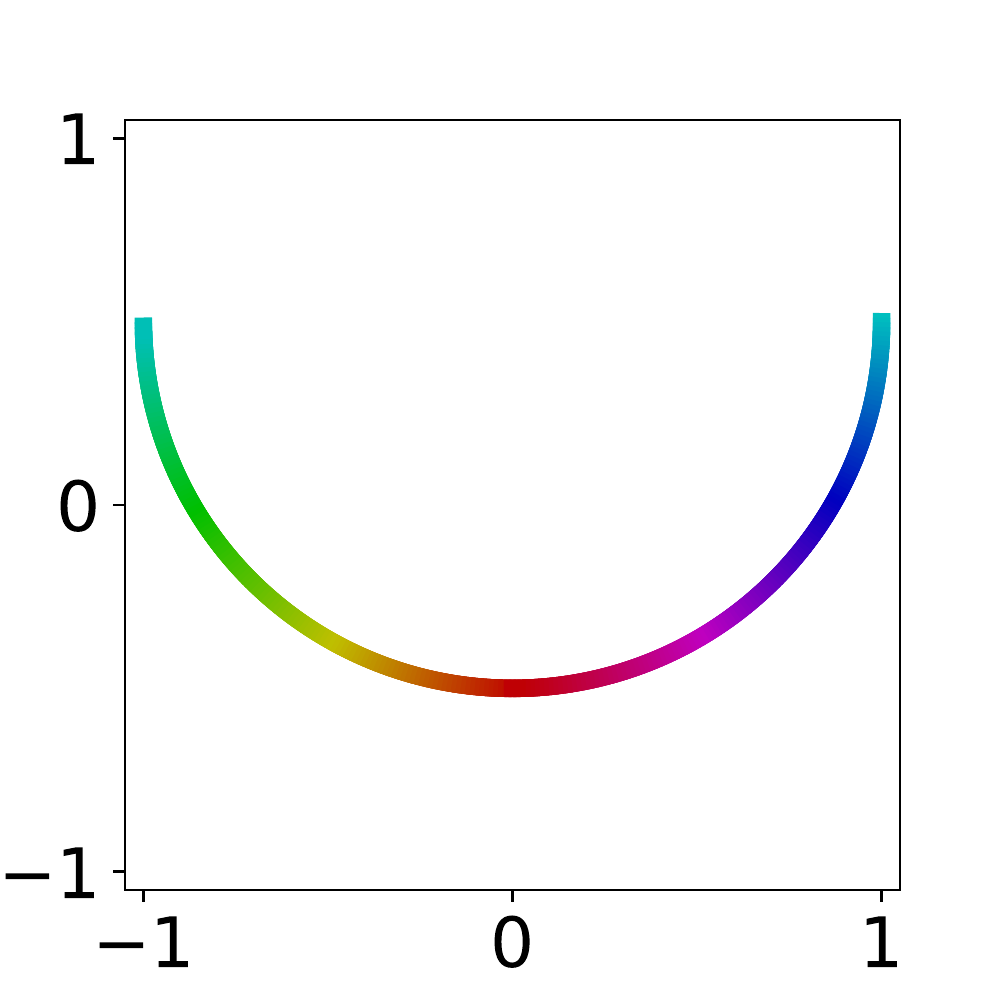} &
        \includegraphics[width=0.93in, trim=0.4in 0.4in 0.4in 0.8in]{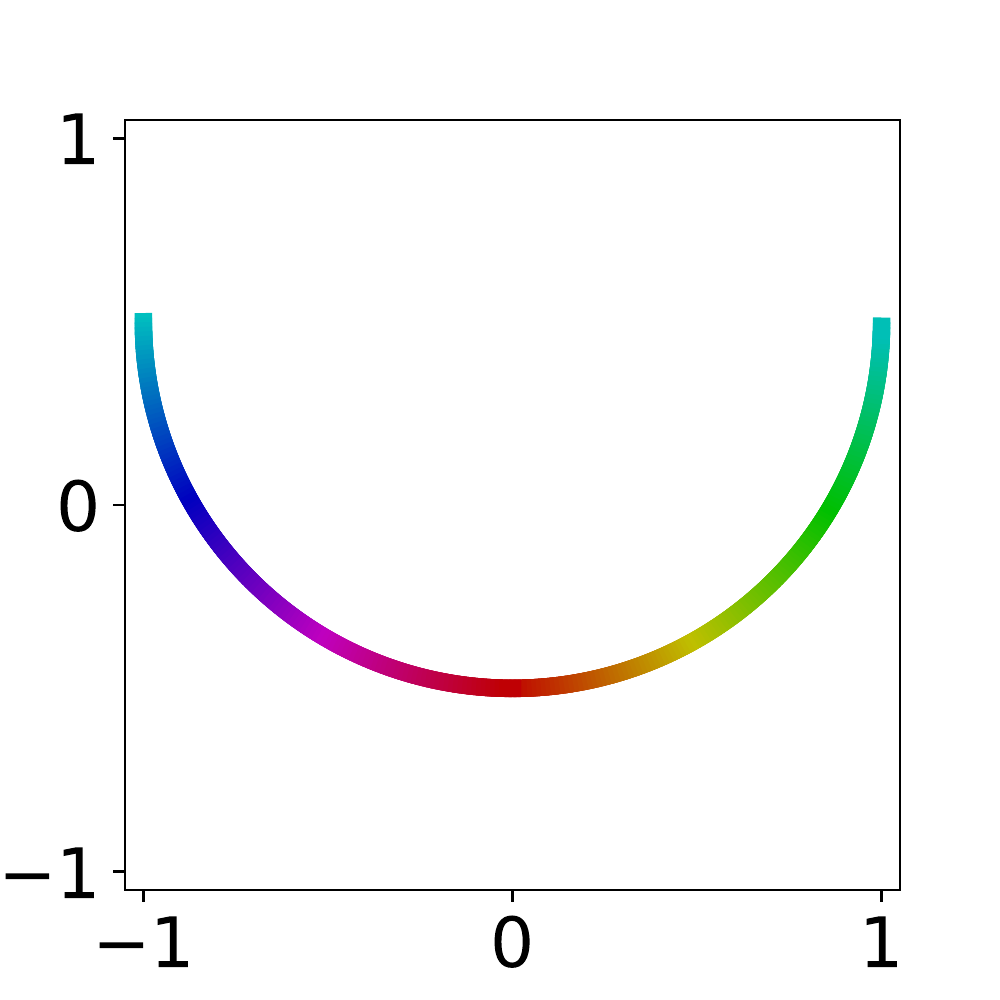} &
        \includegraphics[width=0.93in, trim=0.4in 0.4in 0.4in 0.8in]{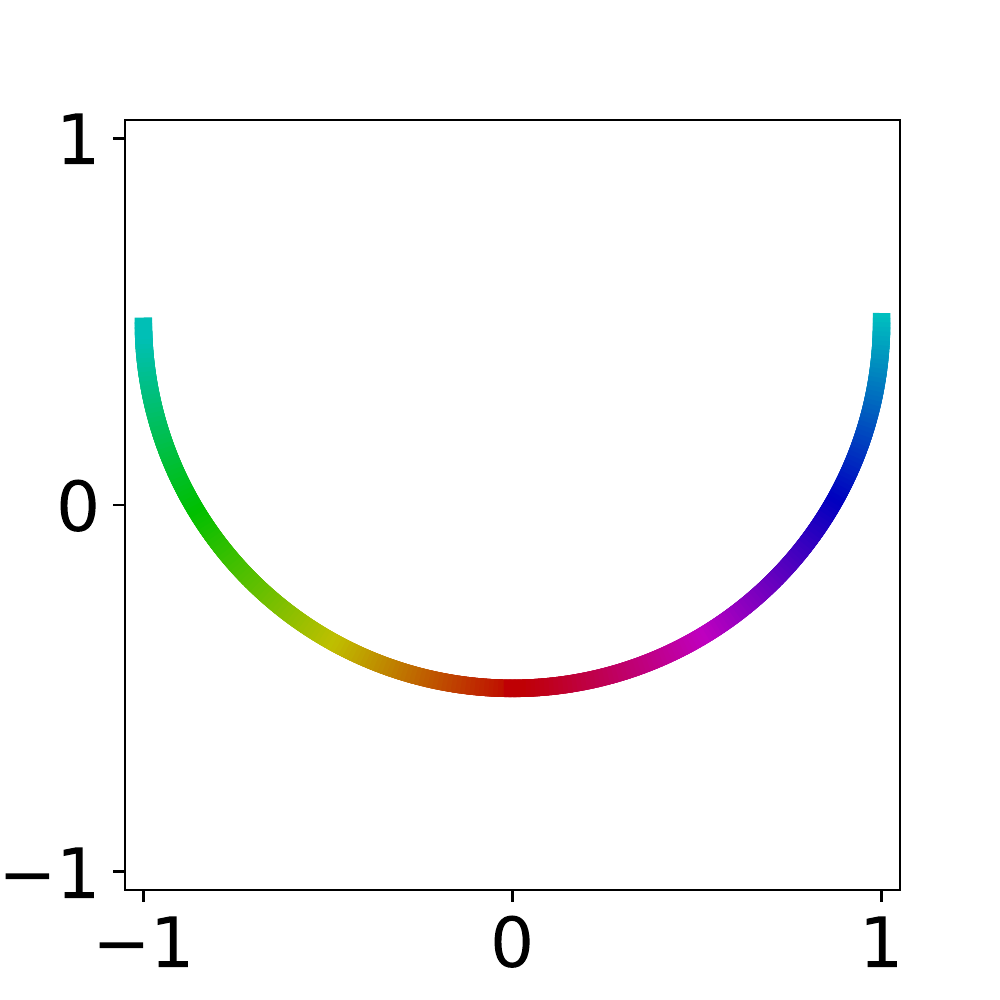} \vspace{0.2cm} \\
        \multicolumn{3}{c}{Axis-angle} \vspace{0.3cm} \\
        \includegraphics[width=0.93in, trim=0.4in 0.4in 0.4in 0.8in]{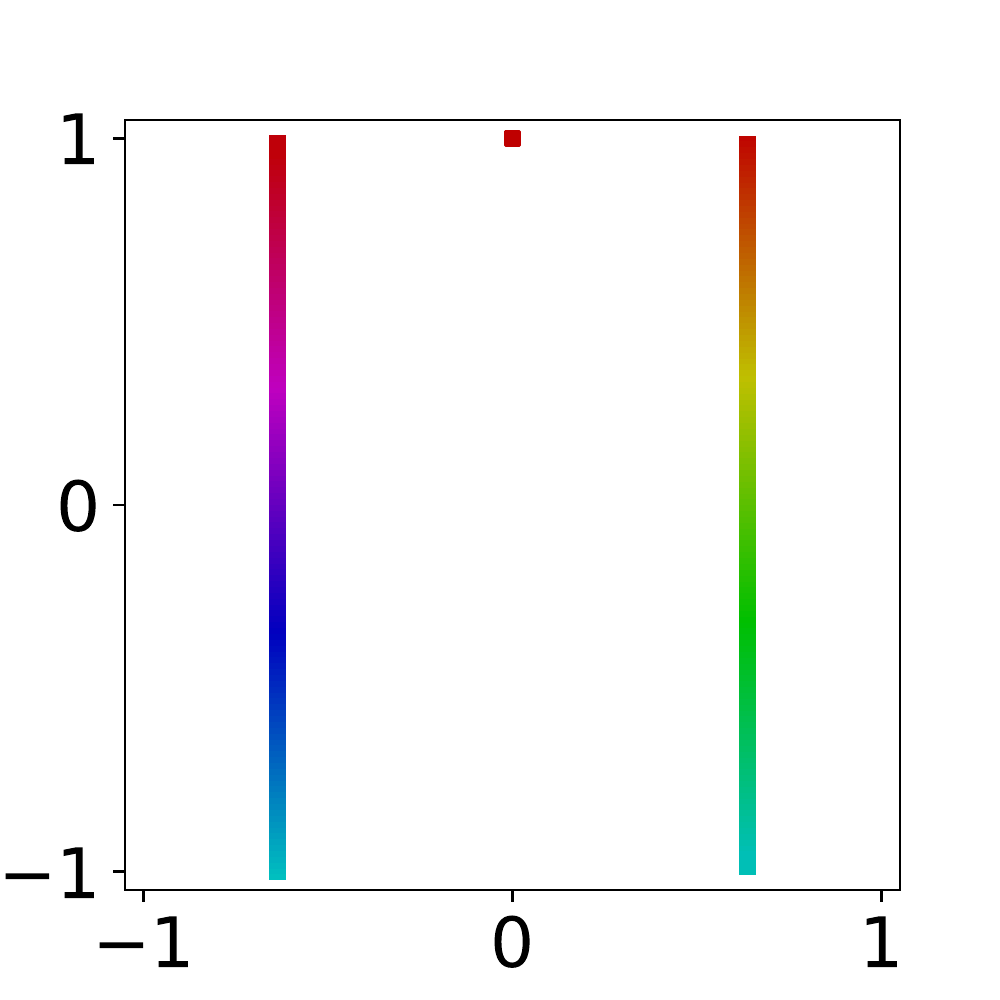} &
        \includegraphics[width=0.93in, trim=0.4in 0.4in 0.4in 0.8in]{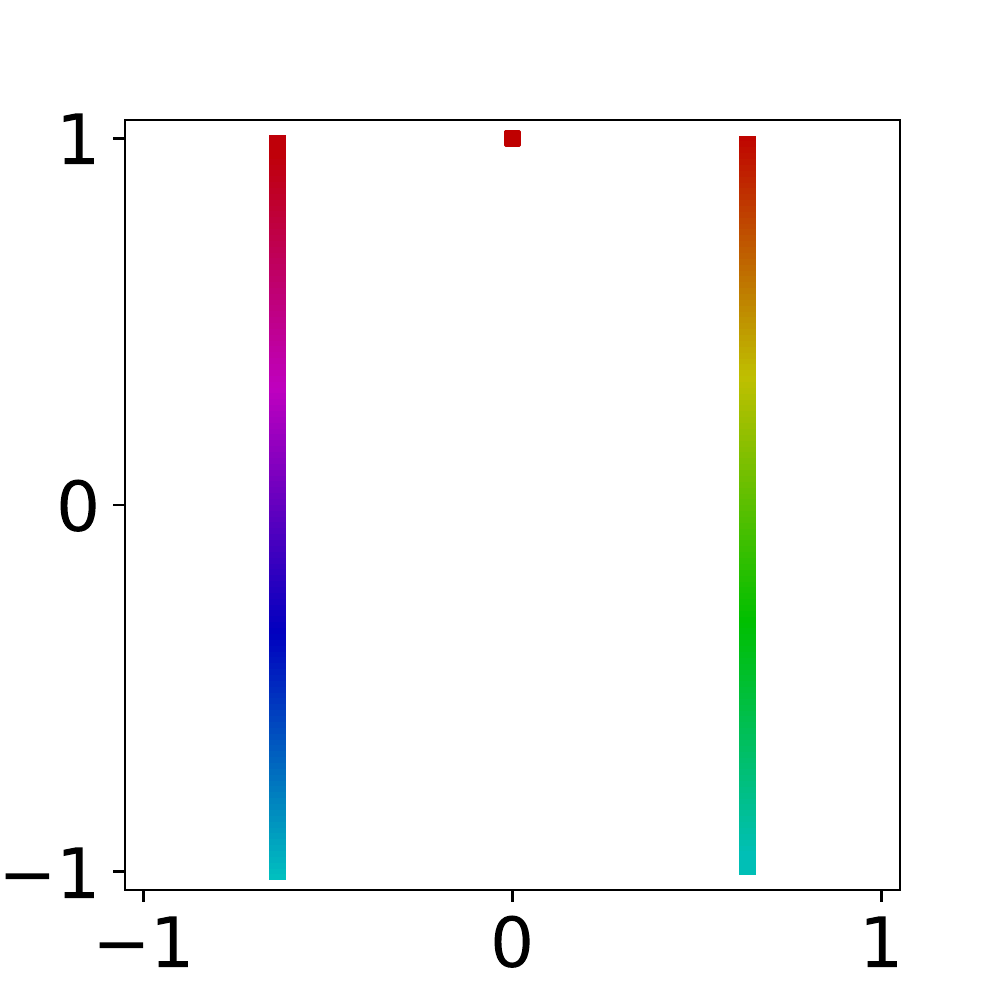} &
        \includegraphics[width=0.93in, trim=0.4in 0.4in 0.4in 0.8in]{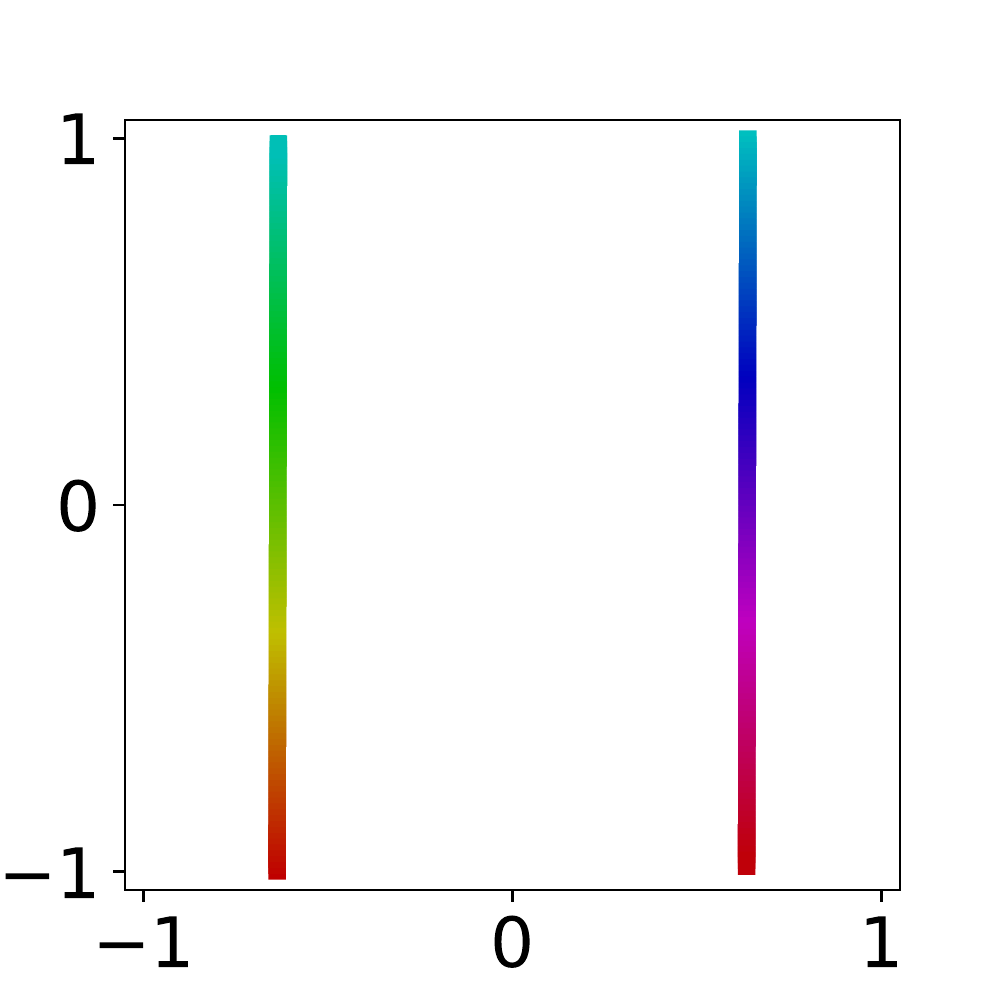} \vspace{0.2cm} \\
        \multicolumn{3}{c}{Euler angles} \vspace{0.3cm} \\
        \includegraphics[width=0.93in, trim=0.4in 0.4in 0.4in 0.8in]{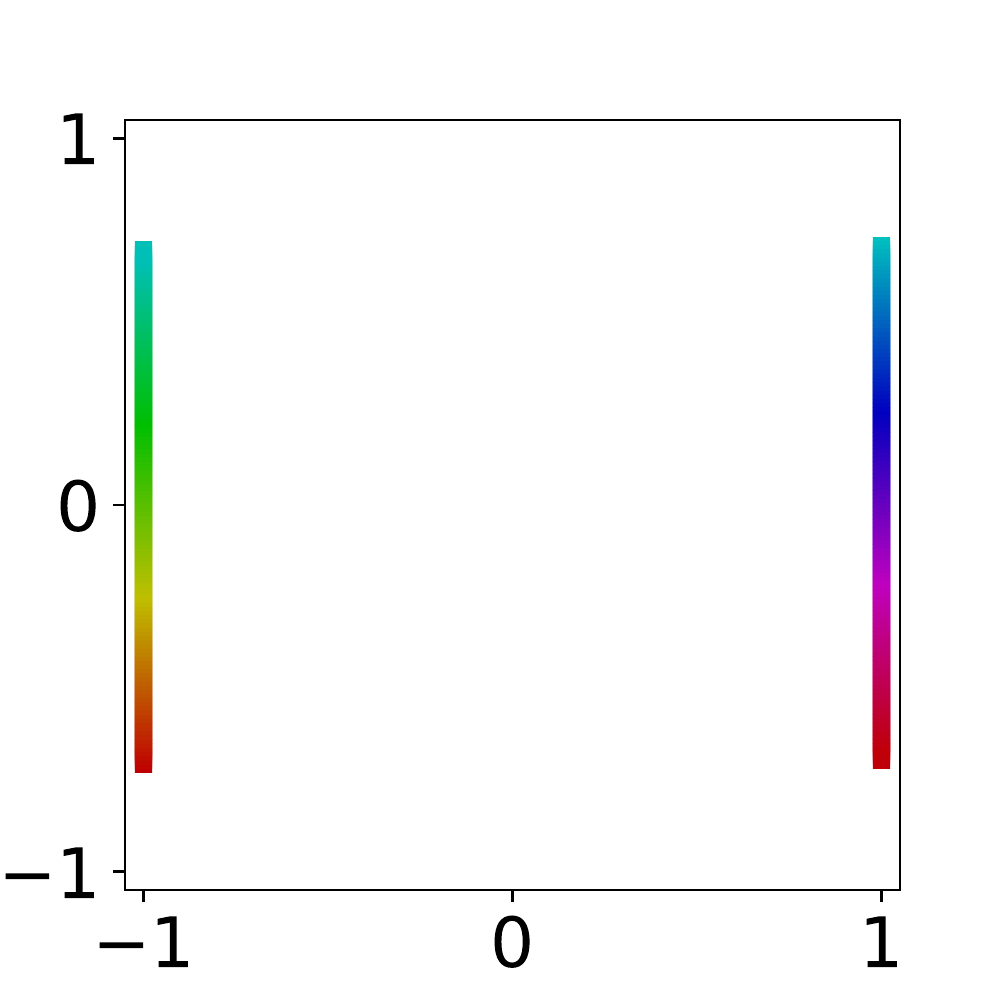} &
        \includegraphics[width=0.93in, trim=0.4in 0.4in 0.4in 0.8in]{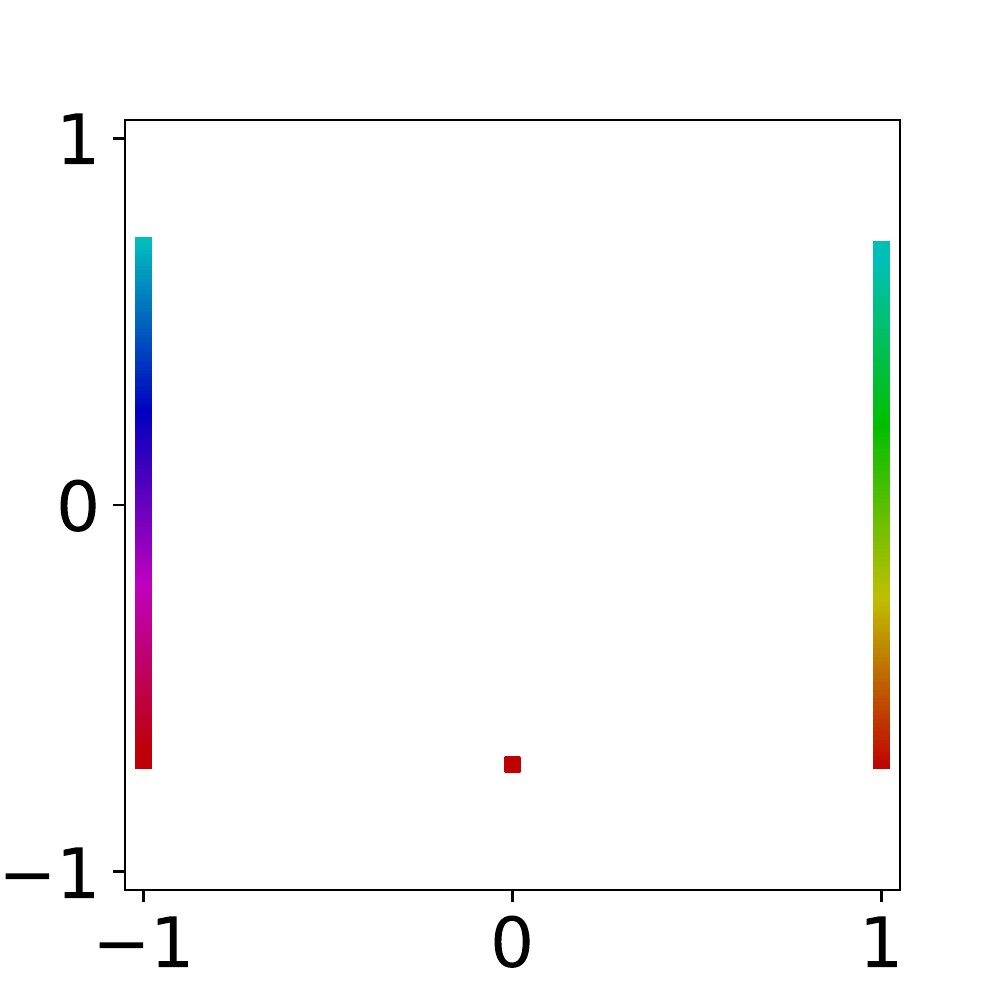} &
        \includegraphics[width=0.93in, trim=0.4in 0.4in 0.4in 0.8in]{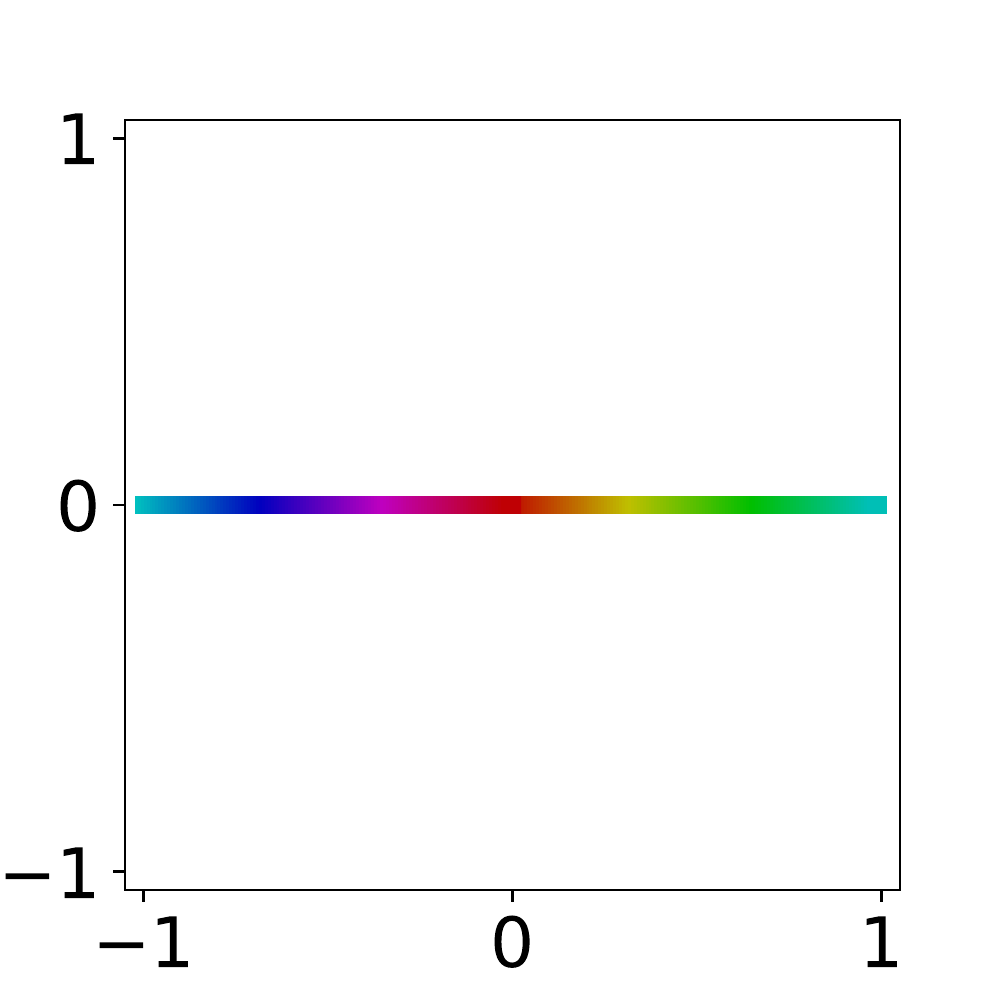} \vspace{0.3cm} \\
        X Rotations & Y Rotations & Z Rotations\vspace{0.1cm} \\
    \end{tabular}
    \caption{Visualization of discontinuities in 3D rotation representations. In the three columns, we show three different curves in $SO(3)$: the ``X, Y, and Z Rotations," which consist of all rotations around the corresponding axis. We map each curve in $SO(3)$ to each of the rotation representations in the different rows (plus the top row, which stays in the original space $SO(3)$), and then map to 2D using PCA. We use the hue to visualize the rotation angle in $SO(3)$ around each of the three canonical axes X, Y, Z. If the representation is continuous then the curve in 2D should be homeomorphic to a circle, and  similar colors should be nearby spatially. We can clearly see that the topology is incorrect for the unit quaternion, axis-angle, and Euler angle representations. }
    \label{fig:vis}
\end{figure}

Here we visualize any discontinuities that might occur in the 3D rotation representations. We do this by forming three continuous curves in $SO(3)$, which we call the ``X, Y, and Z Rotations." We map each of these curves to each representation, and then map the representation curve to 2D by retaining the top two components from Principal Components Analysis (PCA). We call the first curve in $SO(3)$ the ``X Rotations:" this curve is formed by taking the X axis $(1, 0, 0)$, and constructing a curve consisting of all rotations around this axis as parameterized by angle. Likewise, we call the second and third curves in $SO(3)$ the ``Y Rotations" and ``Z Rotations:" these curves are formed by rotating around the Y and Z axes, respectively. We show the resulting 2D curves in \fig{fig:vis}.

\section{Additional Empirical Results}
\label{sec:additional_empirical}
\begin{figure}
    \includegraphics[width=3.2in]{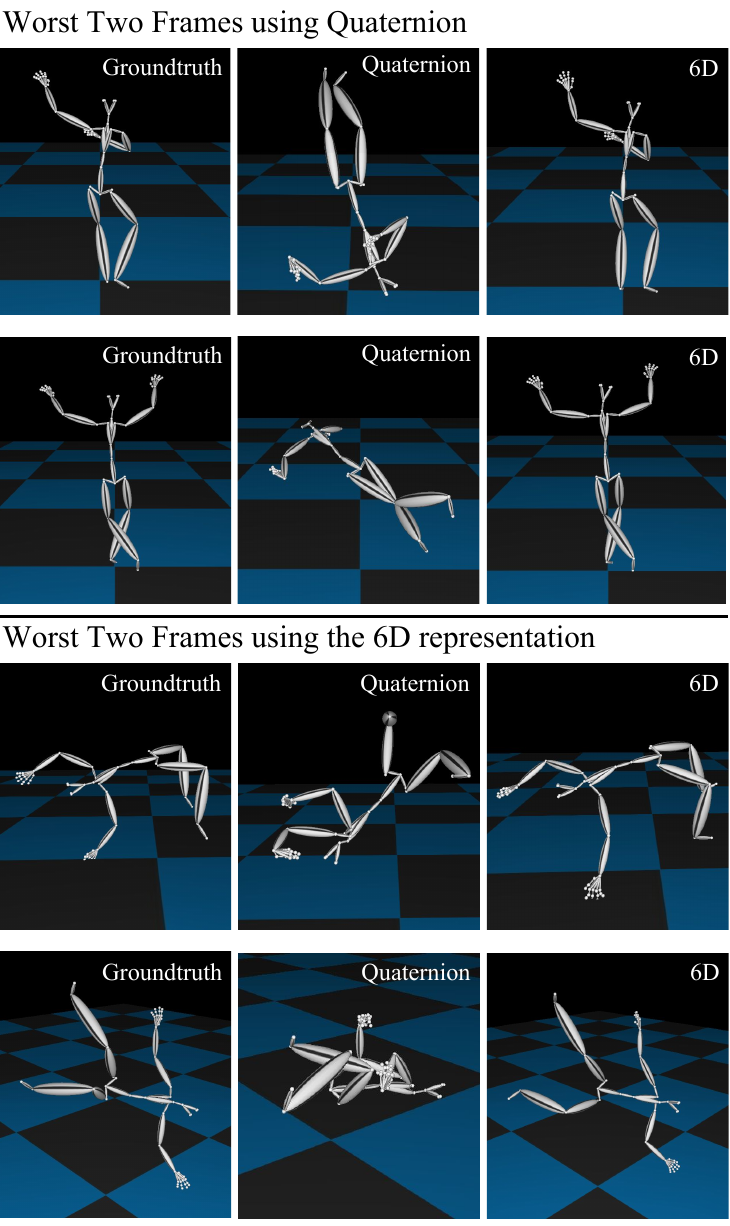}
    \caption{At top, we show IK results for the two frames with highest pose error from the test set for the network trained using quaternions, and the corresponding results on the same frames for the network trained on the 6D representation. At bottom, we show the two worst frames for the 6D representation network, and the corresponding results for the quaternion network.}
    \label{fig:IK_visual_worst}
\end{figure}

In this section, we show some additional empirical results.

\subsection{Visualization of Inverse Kinematics Test Result}
\label{sec:visual_IK}

In \fig{fig:IK_visual_worst}, we visualize the worst two frames with the highest pose errors generated by the network trained on quaternions, along with the corresponding results from the network trained with 6D representations. Likewise, we show the two frames with highest pose errors generated by the network trained on our 6D representation, along with the corresponding results from the network trained on quaternions. This shows that for the worst error frames, the quaternion representation introduces bad qualitative results while the 6D one still creates a pose that is reasonable.

\subsection{Additional Sanity test}
\label{sec:additional_sanity}
\begin{figure*}
    \includegraphics[width=\textwidth]{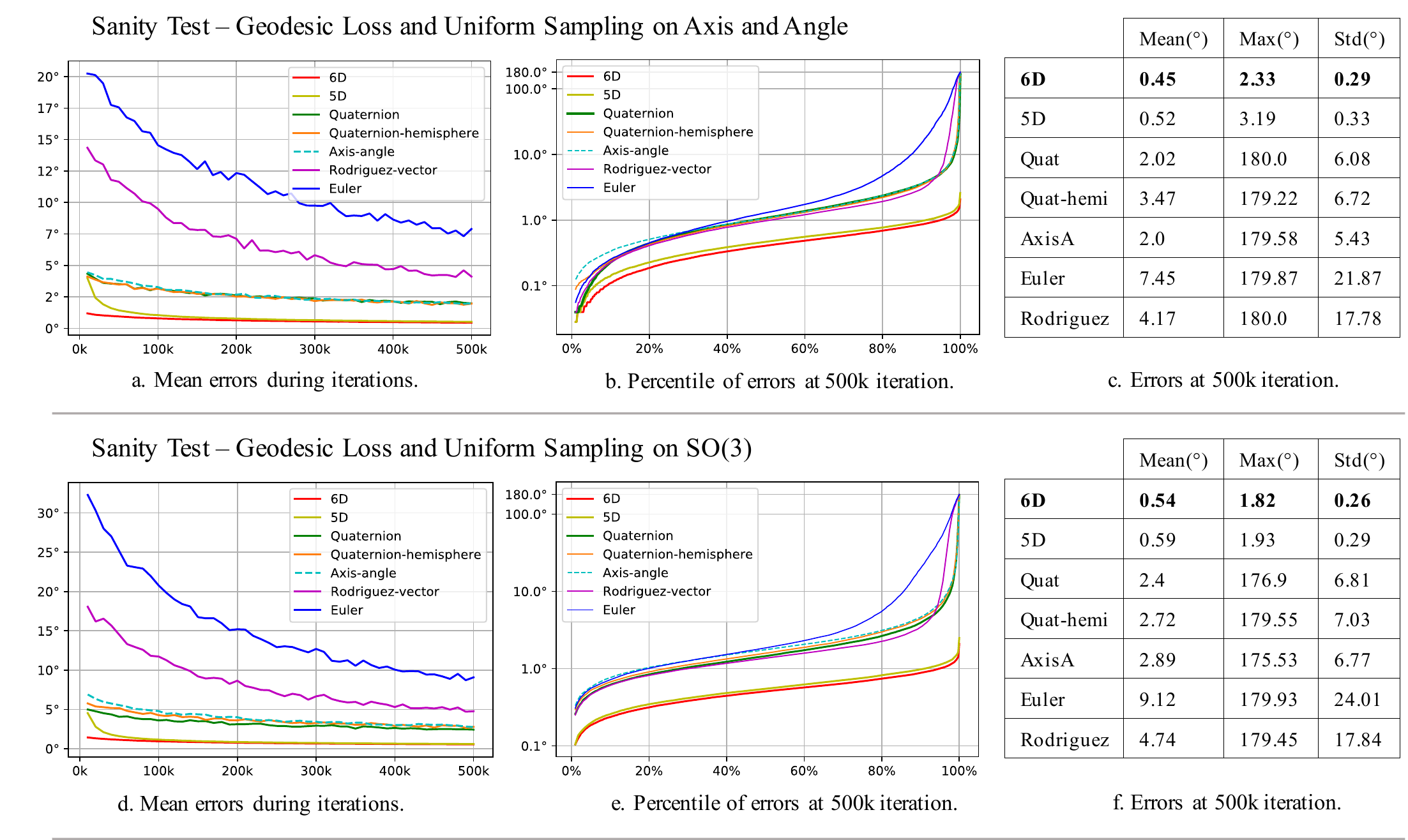}
    \caption{Additional Sanity test results. ``Quat" refers to quaternions, ``Quat-hemi" refers to quaternions constrained to one hemisphere\cite{kendall2017geometric}, ``AxisA" refers to axis angle and ``Rodriguez" refers to the 3D Rodriguez-vector.}
    \label{fig:Sanity_additional}
\end{figure*}

In the main paper, Section \ref{sec:sanity_test}, we show the sanity test result of the network trained with L2 loss between the ground-truth and the output rotation matrices. Another option for training is using the geodesic loss. Besides, the networks in the main paper are trained and tested using a  uniform  sampling  of the axis and the angle which is not a uniform sampling on SO(3) \cite{perez2013uniform}. We present the sanity test result of using the geodesic loss and the two sampling methods in Figure \ref{fig:Sanity_additional}. They are both similar to the result in the main paper.

\textbf{Additional representations.}
In addition to common rotation representations like Euler angles, axis-angles and quaternions, we investigated a few other rotation representations used in recent work including a 3D Rodriguez vector representation, and quaternions that are constrained to one hemisphere as given by Kendall et al.~\cite{kendall2017geometric}. The 3D Rodriguez vector is given as $R=\omega\theta$, where $\omega$ is a 3D unit vector and $\theta$ is the angle~\cite{rodriguez}. We will not provide the proofs for the discontinuity in these representations, but we show their empirical results in Figure \ref{fig:Sanity_additional}. We find that the errors are significantly worse than our 5D and 6D representations.

